\documentclass[review]{elsarticle}

\usepackage[margin=1in]{geometry}
\usepackage{float}
\usepackage{amssymb}
\usepackage{booktabs}
\usepackage{amsmath}
\usepackage{makecell}
\usepackage{mathtools}
\usepackage{subcaption}
\usepackage{lineno}
\usepackage[table]{xcolor}
\usepackage{multirow}
\usepackage{siunitx}
\usepackage{caption}
\usepackage{xparse}
\usepackage{adjustbox}
\usepackage[hidelinks]{hyperref}

\newcommand{\HPGe}{HPGe}
\newcommand{\LaBr}{LaBr$_3$}
\newcommand{\NaI}{NaI(Tl)}
\newcommand{\gadras}{\texttt{GADRAS}}
\newcommand{\geant}{\texttt{Geant4}}
\newcommand{\pyriid}{\texttt{PyRIID}}

\begin{document}

\begin{frontmatter}

\title{Unsupervised domain adaptation for radioisotope identification in gamma spectroscopy}

\author[PNNLaddress]{Peter Lalor\corref{corauthor}}
\author[PNNLaddress,UWaddress]{Ayush Panigrahy}
\author[PNNLaddress]{Alex Hagen}

\cortext[corauthor]{Corresponding author
\\\hspace*{13pt} Email address: peter.lalor@pnnl.gov
\\\hspace*{13pt} Telephone: (925) 453-1876}
\address[PNNLaddress]{Pacific Northwest National Laboratory, Richland, WA 99352 USA}
\address[UWaddress]{University of Washington, Seattle, WA 98195 USA}

\begin{abstract}
Training machine learning models for radioisotope identification using gamma spectroscopy remains an elusive challenge for many practical applications, largely stemming from the difficulty of acquiring and labeling large, diverse experimental datasets. Simulations can mitigate this challenge, but the accuracy of models trained on simulated data can deteriorate substantially when deployed to an out-of-distribution operational environment. In this study, we demonstrate that unsupervised domain adaptation (UDA) can improve the ability of a model trained on synthetic data to generalize to a new testing domain, provided unlabeled data from the target domain is available. Conventional supervised techniques are unable to utilize this data because the absence of isotope labels precludes defining a supervised classification loss. We compare a range of different UDA techniques, finding that feature alignment strategies, particularly via maximum mean discrepancy (MMD) minimization or domain-adversarial training, yield the most consistent improvement to testing scores. For instance, using a custom transformer-based neural network, we achieve a testing accuracy of $0.904 \pm 0.022$ on an experimental \LaBr{} test set after performing unsupervised feature alignment via MMD minimization, compared to $0.754 \pm 0.014$ before alignment. Overall, our results highlight the potential of using UDA to adapt a radioisotope classifier trained on synthetic data for real-world deployment.
\end{abstract}
\end{frontmatter}

\begin{sloppypar}

\section{Introduction}

Improvements to the current performance of gamma spectroscopy are still sought in many applications, especially those with low signal-to-noise ratios (SNRs). Very recent papers cite the need for gamma spectroscopic improvements for mobile source search~\cite{breitenmoserQuantitativeMobileGammaray2026}, nuclear facility upset detection~\cite{fernandezImprovementNuclideDetection2025}, and national security missions~\cite{labovImprovingOperationalPerformance2025a} among others. Many current papers, such as those cited above, focus on improvements to the analysis applied to the data obtained from modern hardware, rather than on improvements to the hardware itself. The abundance of detectable photons from natural surroundings interferes with photons emitted from the source of interest across many situations--a problem which cannot be resolved by different hardware. This problem is analogous to standard problems in computer vision (CV) and natural language processing (NLP), where noise (pixels of unimportant objects in CV or stopwords in NLP) interferes with the signal (pixels of the object of interest in CV or key words conveying meaning in NLP). Unlike in CV and NLP, overparameterized machine learning has thus far evaded widespread adoption in the field of gamma spectroscopy. All three fields have challenges associated with training reliable, generalizable, and explainable models, but the solutions to those challenges from CV and NLP have not been applied to gamma spectroscopy. This is because of the lack of availability of suitable datasets for model training, since collecting and labeling sufficient experimental data is expensive and time-consuming, while relying on synthetic data is inherently limited by the sim-to-real gap~\cite{lalor2025simtoreal}.

Gamma spectroscopy, especially since the advent of portable spectroscopy detectors in the 1950s~\cite{knollRadiationDetectionMeasurement2010}, is one of the most widely used radiation-detection disciplines. The relative ease of detection of photons compared to neutrons, and their relatively high penetration compared to beta and alpha radiation, lead to the ability to perform standoff detection of radiation sources. This standoff detection is useful in radiation safety, national security, and physical science applications. Additionally, the information conveyed by those photons' energy informs the composition of the source, which is useful in material characterization and forensics applications. Gamma spectrometers themselves have differing efficiency, resolution, and portability characteristics; however, they all display a convolution between the energy of the incident photon across the whole spectrum of detected energy. This is irreducible as it originates in the partial deposition of energy during the interaction of photons with matter, and causes challenges in detection and identification scenarios. Further, many natural materials exhibit photon emission, which leads to an ever-present rate of background photon detection events. These two complicating factors may be mitigated through hardware changes, but require advanced analytics for full resolution.

Thus, advanced analytics like machine learning are not new to the field of gamma spectroscopy. Analysis methods have been developed with increasing levels of sophistication, sometimes preceding other fields. For example, automated peak-finding algorithms were proposed as early as 1967~\cite{mariscottiMethodAutomaticIdentification1967} and deployed in commercial software packages by 2006~\cite{Genie2000Operations}; whereas the first algorithm for its computer vision analog, object detection, was not proposed until 2006~\cite{violaRapidObjectDetection2001, zouObjectDetection202023}. Unconstrained optimization-based curve fitting, the mathematical method which forms the basis for modern neural-network-based machine learning, was developed for the analysis of radiation spectra and reached mainstream (physics) software packages more than 30 years ago~\cite{jamesMINUITFunctionMinimization1994}. Only recently, and only in terrestrial applications\footnote{High-energy physics and cosmology have led CV and NLP on analytics development (e.g., graph neural network architectures in track reconstruction using detection events~\cite{reuterEndtoEndMultiTrackReconstruction2024}).}, has gamma spectroscopy lagged CV and NLP on its adoption of cutting-edge machine learning techniques.

Machine learning and its applications to CV and NLP changed dramatically in 2012 with the success of AlexNet for classification of images from the ImageNet dataset~\cite{krizhevskyImageNetClassificationDeep2012}. The key insight leading to that success was that highly over-parameterized functions called neural networks could be optimized on large datasets to minimize the error on prediction tasks. Since then, parameterization has been a key determinant of machine learning success, and in fact ``scaling-laws'' which depend on the number of parameters of the model have since emerged~\cite{hestnessDeepLearningScaling2017, alabdulmohsinRevisitingNeuralScaling2022}. Parameter scaling in gamma spectroscopic analysis has not kept pace with that in CV and NLP. In fact, only some recent publications~\cite{lalor2025simtoreal, ghawaly2022characterization, bandstra2023explaining, bilton2021neural} have embraced the over-parameterized machine learning paradigm.

This slow adoption is likely due to the lack of labeled data in gamma spectroscopy. AlexNet was enabled by the enormous ImageNet dataset, and the aforementioned scaling laws depend on the size of the data as well as the number of parameters in the model. Here again, CV and NLP have developed techniques that could be applicable to gamma spectroscopy, notably domain-adaptation and pretraining. Domain-adaptation has been complementary or competitive with pretraining techniques, largely used to supplement models displaying a generalization gap such as CV models trained on natural imagery transferring to overhead imagery~\cite{farahaniBriefReviewDomain2020}. In CV and NLP, self-supervised or semi-supervised pretraining is used to utilize vast amounts of unlabeled data available from scraped internet sources, and is used in most modern Large Language Models and Vision Language Models~\cite{Vaswani2017, brownLanguageModelsAre2020a, radfordImprovingLanguageUnderstanding, radfordLanguageModelsAre, devlinBERTPretrainingDeep2019}. Especially with the sim-to-real gap identified in gamma spectroscopy~\cite{lalor2025simtoreal}, and the lack of consensus over the use of simulated data in other applied data science fields~\cite{lyImprovingMicrostructuresSegmentation2025, jainScalingLawsLearning2024}, it is important to explore various options for domain adaptation or pretraining-based generalization improvement in gamma spectroscopy. This work presents such a study.

\section{Background}

\subsection{Domain Shift}

Domain adaptation for gamma spectroscopy is formalized as follows: we define $P_s(x, y)$ as the joint data-label probability distribution for a gamma spectrum $x$ with label $y$ for the source domain, and likewise $P_t(x, y)$ for the target domain. Our machine learning objective is to train a neural network classifier $h:\mathcal{X}\to\Delta^K$ to approximate $P_t(y \mid x)$ -- that is, for a given spectrum $x$ from the target domain, we aim to compute the most likely isotopic label $y$.

Lacking labels from the target domain, we might instead train $h$ using labeled source-domain samples to approximate $P_s(y \mid x)$ with the hope that $P_s(y \mid x) \approx P_t(y \mid x)$ and thus the model trained on source-domain data can be used to classify target-domain spectra. However, in practice this is rarely effective due to the \emph{domain shift} between source and target domains, which can be broadly categorized in the following types~\cite{morenotorres2012unifying,
quinonerocandela2009dataset}:

\begin{description}
    \item[Covariate Shift] occurs when the underlying distribution of spectra changes between source and target domain even if the mapping from gamma energy spectrum to isotopic labels remains unchanged. For instance, the target domain might contain more shielded examples than the source domain or fewer spectra with low noise. Formally, pure covariate shift occurs when $P_s(x) \neq P_t(x)$ while $P_s(y \mid x) = P_t(y \mid x)$.

    \item[Prior Shift] occurs when the distribution of isotopic labels differs between source and target domain, even if each isotope produces the same distribution of spectra. For instance, the source domain might contain an equal number of every isotope, whereas the target domain contains a class imbalance. Formally, pure prior shift occurs when $P_s(y) \neq P_t(y)$ while $P_s(x \mid y) = P_t(x \mid y)$.

    \item[Concept Shift] occurs when the fundamental relationship between isotopic label and spectral shape is different between source and target domains. For instance, the source and target domain could have a different geometry, background, gain drift, or detector response. Formally, concept shift occurs when $P_s(x \mid y) \neq P_t(x \mid y)$. This is the most difficult to fix using UDA techniques, since label-agnostic feature alignment typically requires that concept shift is small.
\end{description}

\subsection{Model Architecture}
\label{model architecture}

Past literature in gamma spectroscopy implements a range of different machine learning architectures, including multilayer perceptrons (MLPs)~\cite{Olmos1991, Olmos1992, kamuda2020comparison} and convolutional neural networks (CNNs)~\cite{Liang2019rapid, Moore2019application, moore2020transfer, daniel2020automatic, Barradas2025} to more recent transformer-based neural networks (TBNNs)~\cite{Li2024, lalor2025simtoreal}. We include a summary of each architecture below:

\begin{description}
    \item[Multilayer Perceptrons] represent a gamma spectrum as a 1D list of features and contain one or more hidden layers. In each layer, the features undergo a dense linear transformation followed by a nonlinear activation. MLPs are popular due to their simplicity and flexibility, but lack explicit spatial inductive bias. As a result, MLPs might struggle to learn local structure and are more likely to generalize poorly to spectra with different energy calibrations.

    \item[Convolutional Neural Networks] implement a sliding 1D convolutional kernel across a gamma spectrum, calculating the dot product between the kernel weights and the local spectrum patch. This process captures spatial structure more efficiently than MLPs, but the locality is limited to nearby energy bins, meaning CNNs can be less effective at capturing long-range correlations across a gamma spectrum.

    \item[Transformer-based Neural Networks] leverage the attention mechanism, first introduced in Ref.~\cite{Vaswani2017}, to compute weights between different regions, or ``patches,'' of a spectrum. This enables TBNNs to capture long range correlations in a gamma spectrum, such as multiple decays from different nuclear energy levels. However, transformers are more complex to tune and suffer from quadratic scaling with number of spectral patches.
    
    Transformers were first applied to gamma spectroscopy by Ref.~\cite{Li2024}, henceforth TBNN (Li \emph{et al.}). More recent work introduced a more general transformer architecture with learnable patch embeddings and a [CLS] token, among other enhancements~\cite{lalor2025simtoreal}. In the present work, we build upon this framework with TBNN-LinEmb and TBNN-NonlinEmb, comparing a single affine projection (i.e., one fully connected layer) to a one-hidden-layer MLP embedder. Furthermore, we use a global average pooling readout instead of a learnable [CLS] token, better matching spectrum-level classification by aggregating evidence across all channels.
    
\end{description}

Beyond these architectures, past research has also explored autoencoder-based anomaly detection~\cite{bilton2021neural, ghawaly2022characterization}, recurrent neural networks for source identification~\cite{bilton2021neural}, and custom semi-supervised networks for radioisotope proportion estimation and out-of-distribution identification~\cite{vanomen2024multilabel}. Given the more specialized nature of these methods, we focus the architectural analysis of the present work to MLPs, CNNs, and TBNNs.

\subsection{Unsupervised Domain Adaptation Methods}
\label{unsupervised domain adaptation methods}

In this analysis, we consider the setting where we have access to a labeled \emph{source domain} and an unlabeled \emph{target domain}. The source domain $\mathcal{D}_s=\{(x_i^s,y_i^s)\}_{i=1}^{n_s}$ consists of $n_s$ labeled spectra drawn from $P_s(x,y)$, and the target domain $\mathcal{D}_t=\{x_j^t\}_{j=1}^{n_t}$ consists of $n_t$ spectra drawn from $P_t(x)$ for which labels are not available during training. We first pretrain a \emph{source-only} model using the labeled source domain $\mathcal{D}_s$, writing our model as the composition $h_{\theta,\phi}=g_\phi\circ f_\theta$, where $f_\theta:\mathcal{X}\to\mathbb{R}^d$ is the feature extractor and $g_\phi:\mathbb{R}^d\to\Delta^K$ is the classifier. Here, we represent the feature extractor as the entire neural network up to (and including) the penultimate layer, and the classifier as the final dense classification head. We thus write the source-only pretraining objective as follows:
\begin{equation}
\min_{\theta,\phi}\ \mathcal{L}_{\mathrm{src}}(\theta,\phi)
=\frac{1}{n_s}\sum_{i=1}^{n_s}\ell \left(g_{\phi}(f_{\theta}(x_i^s)),y_i^s\right)~,
\label{eq:source_only}
\end{equation}
where $\ell(\cdot,\cdot)$ is the cross-entropy loss. After pretraining a source-only model by minimizing Eq.~\ref{eq:source_only}, we consider various UDA methods (Sections~\ref{ADDA}--\ref{SimCLR}) which introduce a modified training objective to encourage alignment between feature distributions $f_{\theta}(\mathcal{D}_s)$ and $f_{\theta}(\mathcal{D}_t)$ using unlabeled target samples from $\mathcal{D}_t$. All of the following UDA algorithms were implemented from scratch using TensorFlow~\cite{tensorflow2015whitepaper}.

\subsubsection{Adversarial Discriminative Domain Adaptation (ADDA)}
\label{ADDA}

ADDA~\cite{tzeng2017adversarial} trains a new feature extractor $f_{\theta_t}$ to produce target-domain features which match the source feature distribution (i.e., $f_{\theta_t}(\mathcal{D}_t) \approx f_{\theta}(\mathcal{D}_s)$). This is accomplished by fixing the source feature extractor and adversarially training a target feature extractor $f_{\theta_t}$ and domain discriminator $d_{\psi}:\mathbb{R}^d\to(0,1)$. The domain discriminator's objective is to identify whether a feature vector originated from a source-domain spectrum (0) or a target-domain spectrum (1), trained via the following:

\begin{equation}
\max_{\psi}\ 
\mathbb{E}_{x_s\sim\mathcal{D}_s}\big[\log (1-d_{\psi}(f_{\theta}(x_s)))\big]
+
\mathbb{E}_{x_t\sim\mathcal{D}_t}\big[\log d_{\psi}(f_{\theta_t}(x_t))\big]~,
\label{eq:adda_disc}
\end{equation}
while the target feature extractor is trained to fool the discriminator:
\begin{equation}
\min_{\theta_t}\ 
\mathbb{E}_{x_t\sim\mathcal{D}_t}\big[-\log (1-d_{\psi}(f_{\theta_t}(x_t)))\big]~.
\label{eq:adda_enc}
\end{equation}
The final model is then written as the composition of the source classifier and the target feature extractor: $g_{\phi}\circ f_{\theta_t}$. This approach preserves the source-only model decision boundary while transforming target features to better align with the source domain.

\subsubsection{Deep Adaptation Networks (DAN)}
\label{DAN}

DAN~\cite{long2015learning} adds a term to the supervised source-only loss function to minimize the maximum mean discrepancy (MMD) between the source and target feature vectors $z_s = f_\theta(x_s)$ and $z_t = f_\theta(x_t)$. MMD is defined as the distance between the source and target feature embeddings in a reproducing kernel Hilbert space (RKHS), calculated using the kernel trick:

\begin{equation}
\mathrm{MMD}^2(z_s, z_t)=\left\|
\frac{1}{n_s}\sum_{i=1}^{n_s}\varphi(z_s^i)
-
\frac{1}{n_t}\sum_{j=1}^{n_t}\varphi(z_t^j)
\right\|_{\mathcal{H}}^2
=\frac{1}{n_s^2} \sum_{i,i'}k(z_s^i,z_s^{i'})
+\frac{1}{n_t^2} \sum_{j,j'}k(z_t^j,z_t^{j'})
-\frac{2}{n_sn_t} \sum_{i,j}k(z_s^i,z_t^j)~,
\label{eq:mmd}
\end{equation}

where $k$ is a positive definite kernel and $\varphi$ is its feature map. The DAN training objective is thus 

\begin{equation}
\min_{\theta,\phi}\ \mathcal{L}_{\mathrm{src}}(\theta,\phi)
+\lambda \mathrm{MMD}^2(z_s, z_t)~,
\label{eq:dan_obj}
\end{equation}

where $\lambda \geq 0$ is a tradeoff parameter which balances the contribution of the source-only supervised term with the MMD feature alignment penalty. In our experiments, we compute MMD using a multi-kernel Gaussian radial basis function (RBF), averaging over several kernel bandwidths and omitting diagonal terms. The resulting domain-adapted model thus retains the ability to correctly classify source-domain spectra while simultaneously producing aligned source and target feature vectors, yielding better generalization to the target domain.

\subsubsection{Domain Adversarial Neural Networks (DANN)}
\label{DANN}

DANN~\cite{ganin2016domain} trains the source feature extractor to learn domain-invariant features. This is accomplished by training a domain discriminator $d_{\psi}:\mathbb{R}^d\to(0,1)$ while adversarially updating the feature extractor to fool the discriminator. The min--max objective is thus:
\begin{equation}
\min_{\theta,\phi}\ \max_{\psi}\ 
\mathcal{L}_{\mathrm{src}}(\theta,\phi)
-\kappa 
\mathbb{E}_{x\sim(\mathcal{D}_s\cup\mathcal{D}_t)}
\Big[\ell_d\big(d_{\psi}(f_{\theta}(x)),r(x)\big)\Big]~,
\label{eq:dann_obj}
\end{equation}

where $\kappa \geq 0$ is the domain-adversarial strength, $r(x)\in\{0,1\}$ is the true domain label (0=source, 1=target), and $\ell_d(\cdot, \cdot)$ is the binary cross-entropy loss. In practice, Eq.~\ref{eq:dann_obj} is optimized using a gradient reversal layer that multiplies the gradient from the discriminator loss entering the feature extractor by $-\kappa$ during backpropagation. After domain adaptation, the feature extractor produces a domain-agnostic representation and is thus less likely to overfit to features only present in the source domain.

\subsubsection{Deep Correlation Alignment (DeepCORAL)}
\label{DeepCORAL}

DeepCORAL~\cite{sun2016deepcoral} aligns the second-order statistics of the source and target distributions by minimizing the difference between their covariance matrices. Defining covariances $C_s$ and $C_t$ for minibatch feature matrices $Z_s\in\mathbb{R}^{n_s\times d}$ and $Z_t\in\mathbb{R}^{n_t\times d}$ as:
\begin{equation}
C_s=\frac{1}{n_s-1}(Z_s-\mathbf{1}\bar z_s^\top)^\top(Z_s-\mathbf{1}\bar z_s^\top)~,
\qquad
C_t=\frac{1}{n_t-1}(Z_t-\mathbf{1}\bar z_t^\top)^\top(Z_t-\mathbf{1}\bar z_t^\top)~,
\label{eq:coral_cov}
\end{equation}
we express the CORAL loss as:
\begin{equation}
\mathcal{L}_{\mathrm{CORAL}}=\frac{1}{4d^2}\|C_s-C_t\|_F^2~,
\label{eq:coral_loss}
\end{equation}
and subsequently optimize the network via the combined objective:
\begin{equation}
\min_{\theta,\phi}\ \mathcal{L}_{\mathrm{src}}(\theta,\phi)+\lambda \mathcal{L}_{\mathrm{CORAL}}~,
\end{equation}
where $\lambda$ is a tradeoff parameter. The result is a simple and lightweight way to align source and target feature distributions.

\subsubsection{Deep Joint Optimal Transport (DeepJDOT)}
\label{DeepJDOT}

DeepJDOT~\cite{damodaran2018deepjdot} aligns the \emph{joint} distributions of features and labels using optimal transport (OT). For each minibatch, a pairwise cost is defined as

\begin{equation}
c_{ij}(\theta,\phi)=
\alpha \|f_{\theta}(x_s^i)-f_{\theta}(x_t^j)\|_2^2
+\beta \ell \left(g_{\phi}(f_{\theta}(x_t^j)),y_s^i\right),
\label{eq:deepjdot_cost}
\end{equation}

where $\ell(\cdot,\cdot)$ is the cross-entropy loss. The first term in Eq.~\ref{eq:deepjdot_cost} measures the similarity between samples $i$ and $j$ in feature space, and the second term measures the agreement in label space between the \textit{true} source label $y_s$ and the \textit{predicted} target label $\hat y_t = g_{\phi}(f_{\theta}(x_t^j))$. In Eq.~\ref{eq:deepjdot_cost}, $\alpha, \beta \geq 0$ are hyperparameters which balance the feature-distance and label-consistency. We then compute a \emph{soft matching} (transport plan) $\gamma\in\mathbb{R}_+^{n_s\times n_t}$ between the $n_s$ source spectra and $n_t$ target spectra by solving the entropic OT problem with uniform marginals using Sinkhorn iterations. We subsequently define the OT loss as follows:
\begin{equation}
\mathcal{L}_{\mathrm{OT}}(\theta,\phi)=\sum_{i=1}^{n_s}\sum_{j=1}^{n_t}\gamma_{ij} c_{ij}(\theta,\phi)~.
\label{eq:deepjdot_otloss}
\end{equation}
Our training objective is thus:
\begin{equation}
\min_{\theta,\phi}\ \mathcal{L}_{\mathrm{src}}(\theta,\phi)+\lambda \mathcal{L}_{\mathrm{OT}}(\theta,\phi)~,
\label{eq:deepjdot_obj}
\end{equation}
for a tradeoff parameter $\lambda$. Through this OT procedure, we achieve a class aware method of aligning source and target domains.

\subsubsection{Mean Teacher}
\label{Mean Teacher}

Mean Teacher~\cite{tarvainen2018meanteachersbetterrole} enforces predictive consistency under noise/augmentations between a \emph{teacher} model and \emph{student} model, where the weights of the teacher are an exponential moving average (EMA) of the student's weights. For each target spectrum $x_t$, two stochastic views $x_t^{(s)}$ and $x_t^{(t)}$ are formed by adding Poisson noise to the original spectrum, with an \emph{effective counts} hyperparameter introduced to control the scale of the noise\footnote{Alternative data augmentations are possible, such as background addition, detector broadening, peak masking, or gain shift~\cite{stomps2023data}. We selected Poisson resampling to serve as a simple baseline.} We then define a self-supervised consistency loss as the mean squared error between the student's predictions $g_{\phi_s}(f_{\theta_s}(x_t^{(s)}))$ and the teacher's predictions $g_{\phi_t}(f_{\theta_t}(x_t^{(t)}))$:

\begin{equation}
\mathcal{L}_{\mathrm{cons}}(\theta_s,\phi_s;\theta_t,\phi_t)
=
\mathbb{E}_{x_t\sim \mathcal{D}_t}\left[
\left\|g_{\phi_s}(f_{\theta_s}(x_t^{(s)})) - g_{\phi_t}(f_{\theta_t}(x_t^{(t)}))\right\|_2^2
\right]~.
\end{equation}

The student parameters are optimized via the combined training objective:

\begin{equation}
\min_{\theta_s,\phi_s}\ 
\mathcal{L}_{\mathrm{src}}(\theta_s,\phi_s)
+\lambda\,\mathcal{L}_{\mathrm{cons}}(\theta_s,\phi_s;\theta_t,\phi_t)~.
\end{equation}

for a tradeoff parameter $\lambda$, and the teacher parameters are updated by EMA:

\begin{equation}
\theta_t \leftarrow \mu \theta_t + (1-\mu)\theta_s~,
\qquad
\phi_t \leftarrow \mu \phi_t + (1-\mu)\phi_s,
\label{eq:meanteacher_ema}
\end{equation}

where $\mu\in[0,1)$ is the EMA decay. Mean Teacher encourages stable predictions on target spectra while remaining robust to noise.

\subsubsection{Simple Contrastive Learning (SimCLR)}
\label{SimCLR}

SimCLR~\cite{chen2020simpleframeworkcontrastivelearning} performs self-supervised learning by contrasting two stochastic views of the same underlying spectrum. Similar to Mean Teacher, we perform independent Poisson resampling on each target spectrum $x_t$ to generate two views $x_t^{(1)}$ and $x_t^{(2)}$. We subsequently introduce a projection head $q_\omega:\mathbb{R}^d\to\mathbb{R}^{d'}$ and compute $z_1 = q_\omega(f_\theta(x_t^{(1)}))~,~z_2 = q_\omega(f_\theta(x_t^{(2)}))$. For a minibatch of $m$ target spectra, we define a contrastive InfoNCE loss as:

\begin{equation}
\mathcal{L}_{\mathrm{NCE}}=\frac{1}{2}\left(\mathcal{L}_{1\rightarrow 2}+\mathcal{L}_{2\rightarrow 1}\right)~,\qquad
\mathcal{L}_{1\rightarrow 2}
=
\frac{1}{m}\sum_{i=1}^{m}
-\log
\frac{\exp\left(\mathrm{sim}(z_1^i,z_2^i)/\tau\right)}
{\sum_{j=1}^{m}\exp\left(\mathrm{sim}(z_1^i,z_2^j)/\tau\right)}
\end{equation}

where $\mathrm{sim}(u,v) = \frac{u^T v}{\|u\| \|v\|}$ is the cosine similarity and $\tau > 0$ is a temperature hyperparameter. The overall training objective is thus:

\begin{equation}
\min_{\theta,\phi,\omega} \mathcal{L}_{\mathrm{src}}(\theta,\phi) + \lambda\,\mathcal{L}_{\mathrm{NCE}}(\theta,\omega)~,
\end{equation}

where $\lambda \geq 0$ is a tradeoff parameter. SimCLR encourages a model to learn augmentation-invariant representations of target spectra while retaining the ability to correctly classify source spectra.

\section{Methodology}

\subsection{Dataset Curation}
\label{dataset curation}

In this analysis, we consider three domain adaptation scenarios, summarized in Table~\ref{table:datasets}. In the first domain adaptation scenario, \emph{sim-to-sim}, our target domain is simulated in \geant{}, an open-source Monte Carlo simulation package for modeling the passage of radiation through matter~\cite{allison2016recent, archambault2023g4ares}. We simulated high-resolution spectra for 55 radioisotopes, placing each in a cylindrical sample container, and measuring incoming gamma rays using a high-purity germanium (\HPGe) detector placed one meter from the source, mirroring the geometry from Ref.~\cite{pierson2022alpha}. Five representative detector geometries and five source-container configurations (aluminum, thorium, cellulose nitrate, water, and copper) were used to capture a range of realistic counting scenarios. These template spectra are then mixed, background-added, and Poisson resampled using \pyriid{}, a Python package for gamma spectral synthesis~\cite{Morrow2021PyRIID}. The resulting dataset is class-balanced and comprises exclusively solid isotopes, with SNRs log-uniformly sampled between 100 and 1000. Each spectrum contains up to 14 isotopes, with relative source contributions drawn from a Dirichlet distribution ($\alpha=3$).

The second and third domain adaptation scenarios, \emph{sim-to-real}, reflect a different measurement scenario from the sim-to-sim simulations. Our target domain consists of experimental spectra measured using handheld sodium iodide (\NaI) and lanthanum bromide (\LaBr) detectors, respectively, measured in standardized source-shield configurations (none, steel, lead, polyethylene). The spectra were interpolated onto a uniform 1024-bin energy grid from 0 to 3000~keV, and each source belonged to one of 32 isotope classes (partially overlapping the sim-to-sim isotopes). Due to data sensitivity constraints, detailed information about the sim-to-real experimental datasets (e.g., specific isotope lists or acquisition protocols) is not available for public release.

For each domain adaptation scenario, we also simulated a corresponding source-domain dataset using \gadras{}. \gadras{} is a semi-empirical detector response calculation software which couples deterministic attenuation calculations with precomputed response functions to quickly generate synthetic template spectra~\cite{horne2016gadras}. These high resolution template spectra are then mixed (sim-to-sim only), background-added, and Poisson resampled using \pyriid{} to efficiently generate a large dataset for machine learning training. For the sim-to-sim dataset, the SNR was sampled log-uniformly between 100 and 1000. For the sim-to-real datasets, the SNR was instead drawn from an exponential distribution, with a mean of 896 for LaBr and 1370 for NaI. All source-domain datasets were class-balanced, and the dataset curation process is described in further detail in our previous research~\cite{lalor2025simtoreal}.

The datasets were partitioned into training, validation, and testing with a 70/15/15 split. Splits were performed randomly rather than stratified by class, and all reported results correspond to a single fixed split. In all three domain adaptation scenarios, the isotopic labels of the target domain spectra were masked to imitate a setting where ground truth composition is unknown. The isotopic labels of the validation and testing sets were not masked in order to evaluate the performance of trained models for algorithmic comparison.

\begin{table}
\centering
\caption{Summary of the datasets used in this study.}
\begin{tabular}{@{}lllccc@{}}
\toprule
\textbf{Scenario} 
  & \textbf{Domain} 
  & \makecell{\textbf{Acquisition}\\[-0.5ex]\textbf{Mode}} 
  & \makecell{\textbf{Isotopic}\\[-0.5ex]\textbf{Composition}} 
  & \makecell{\textbf{Number of}\\[-0.5ex]\textbf{Isotopes}} 
  & \textbf{Size} \\
\midrule
\multirow{2}{*}{sim-to-sim (\HPGe)}
  & Source & \gadras{}     & Mixed  & 55 & $1.3\times10^6$ \\
  & Target & \geant{}     & Mixed  & 55 & $1.3\times10^6$ \\
\addlinespace
\multirow{2}{*}{sim-to-real (\LaBr)}
  & Source & \gadras{}     & Single & 32 & $1.4\times10^6$ \\
  & Target & Experiment & Single & 32 & 15,091        \\
\addlinespace
\multirow{2}{*}{sim-to-real (\NaI)}
  & Source & \gadras{}     & Single & 32 & $1.4\times10^6$ \\
  & Target & Experiment & Single & 32 & 10,440        \\
\bottomrule
\end{tabular}
\label{table:datasets}
\end{table}

The three scenarios considered in this analysis differ not only in the nature of the domain shift (sim-to-sim versus sim-to-real), but also in task structure. The sim-to-sim scenario contains 55 isotope classes, and each spectrum contains an arbitrary mixture of up to 14 isotopes, while the sim-to-real scenarios involve single-label classification over a partially overlapping subset of 32 isotope classes. These scenarios are thus not intended as direct comparisons against each other, but rather as independent evaluations of UDA across a wide range of settings in gamma spectroscopy. We note that the two sim-to-real scenarios (\LaBr{} and \NaI{}) are well-matched to each other (same 32 isotopes and single-isotope task) and differ only in detector type, so cross-comparison between those two is well-controlled.

\subsection{Model Training}
\label{model training}

As an input preprocessing step, we applied a variance-stabilizing square-root transform to the raw channel counts, $\tilde x = \sqrt{x}$. We subsequently $z$-score normalized the transformed counts on a per-spectrum basis, $\hat x = (\tilde x - \mu) / \sigma$, where $\mu$ and $\sigma$ are the mean and standard deviation of $\tilde x$ computed across channels for a single spectrum. We found this preprocessing improved training stability, especially for the transformer architectures.

To train the source-only models, we first performed a Bayesian hyperparameter search using the \texttt{Optuna} software package~\cite{optuna2019}. For each architecture, we performed 100 Bayesian trials with a validation loss search criterion. For consistency, we used the same architecture hyperparameters across all scenarios (selected as the best source-only parameters from the sim-to-sim dataset), but training hyperparameters (learning rate, batch size, weight decay, dropout) were calculated separately for every run. Once the best hyperparameters were selected, we trained 10 models using different random weight initializations, per-epoch data shuffling, and dropout seeds. This ensemble technique was chosen as a simple form of uncertainty estimation, as the $1\sigma$ sample standard deviation of metric evaluations across the 10 randomized trials quantifies realistic run-to-run performance fluctuations.

After source-only models were trained, we performed an additional 100 Bayesian trials to determine the best UDA-specific hyperparameters. We performed this search independently for each architecture (Section~\ref{model architecture}) and for each UDA method (Section~\ref{unsupervised domain adaptation methods}). Once the best hyperparameters were selected, we again trained 10 models using different random seeds to estimate uncertainty in model performance. The domain-adapted models and source-only models were paired by using the corresponding source-only model as the initial configuration prior to unsupervised domain alignment. We include a summary of the selected hyperparameters for each domain adaptation scenario across all architectures and UDA methods in Tables~\ref{table:source_arch_hps}--\ref{table:SimCLR_hps}.

\section{Results}

\subsection{Domain Shift Characterization}
\label{domain shift characterization}

To illustrate the presence of source-to-target domain shift for each of the domain adaptation scenarios, we computed three domain-gap metrics. First, we trained a linear feature-space domain discriminator $d_{\psi}:\mathbb{R}^d\to(0,1)$ to identify whether a feature vector $z_{s,t} = f_\theta(x_{s,t})$ originated from the source-domain (0) or target-domain (1). We also trained an input-space domain discriminator directly on $x_{s,t}$. We subsequently computed the area under the receiver operating characteristic curve (AUROC) on a held-out test set to quantify how easily each domain discriminator was able to differentiate between source and target domains (AUROC = 0.5 $\implies$ random guessing, AUROC = 1 $\implies$ perfect domain discrimination)~\cite{David2010theory}. We supplemented this analysis by empirically estimating the MMD (Eq.~\ref{eq:mmd}) between source and target $\ell_1$-normalized spectra using an RBF kernel with median heuristic bandwidth ($\mathrm{MMD} = 0 \implies$ identical distributions, larger values indicate larger domain gap)~\cite{Gretton2012kernel}. Finally, we defined a class-conditional expected pairwise Wasserstein distance as the average Wasserstein-1 distance $W_1(\cdot,\cdot)$ between random $\ell_1$-normalized spectra for a given isotopic label $c$~\cite{Peyr2019computational}. We then compute a custom heuristic $\mathrm{R}$ by comparing the source-to-target class-conditional expected pairwise Wasserstein distance to the same distance evaluated within each domain:

\begin{equation}
\mathrm{R}(\mathcal{D}_s,\mathcal{D}_t)
\coloneqq
\frac{\sum_{c=1}^{K} \widetilde{W}_{s,t}^{(c)}}
{\tfrac12\sum_{c=1}^{K}\left(\widetilde{W}_{s,s}^{(c)}+\widetilde{W}_{t,t}^{(c)}\right)}~,
\qquad
\widetilde{W}_{s,t}^{(c)}
\coloneqq
\mathop{\mathbb{E}}\limits_{\substack{
x_s \sim \mathcal{D}_s^{c}\\
x_t \sim \mathcal{D}_t^{c}
}}
\left[W_1\left(x_s, x_t\right)\right]~.
\label{eq:expected_pairwise_W1}
\end{equation}

Intuitively, we interpret the Wasserstein distance $W_1(\cdot,\cdot)$ between two spectra as the minimal transport cost to morph one spectrum into another, and $\mathrm{R}$ as the ratio of between-domain to within-domain spectral dissimilarity. Thus $\mathrm{R}=1 \implies$ consistency between $\mathcal{D}_s$ and $\mathcal{D}_t$, whereas $\mathrm{R} \gg 1 \implies$ substantial dissimilarity. We compute Eq.~\ref{eq:expected_pairwise_W1} in the sim-to-sim scenario using single-isotope (one-hot) spectra from both the source and target datasets, since mixed spectra cannot be uniquely assigned to a single class $c$. For each domain separability metric, we computed a 95\% confidence interval (CI) using a stratified bootstrapping approach, and performed statistical significance testing via an approximate bootstrap for AUC (null = 0.5) and $\mathrm{R}$ (null = 1) and a permutation test for MMD (null = 0). These results are summarized in table~\ref{tab:domain_shift}, where we see that all three metrics identify a statistically significant source-to-target domain gap across all domain adaptation scenarios. This result highlights the importance of leveraging domain adaptation techniques to reduce the impact of domain shift on model generalizability.

\begin{table}
\centering
\begin{tabular}{l c c c c}
\toprule
\textbf{Scenario} & \textbf{AUROC (features)} & \textbf{AUROC (inputs)} & \textbf{MMD$^2$} & 
$\mathrm{R}$ \\
\midrule
sim-to-sim (\HPGe) & [0.993, 0.996] & [0.957, 0.967] & [2.74, 2.93] & [1.36, 1.40] \\
sim-to-real (\LaBr)& [0.982, 0.988] & [0.852, 0.874] & [2.26, 2.55] & [1.87, 1.93]\\
sim-to-real (\NaI) & [0.983, 0.989] & [0.780, 0.808] & [1.30, 2.53] & [1.93, 1.99]\\
\bottomrule
\end{tabular}
\caption{95\% CI for each domain separability metric, indicating a statistically significant domain gap between source and target domains for all domain adaptation scenarios.}
\label{tab:domain_shift}
\end{table}

\subsection{Testing Scores}
\label{testing scores}

The results of this analysis are summarized in Table~\ref{table:UDA_scores}, where we compare the performance of different machine learning models on a held-out testing dataset, with $1\sigma$ uncertainties calculated as the sample standard deviations across the 10 randomized trials described in Section~\ref{model training}. In the sim-to-real scenarios, we compute the accuracy on the held-out experimental testing dataset as our evaluation metric. In the sim-to-sim scenario, we instead use the APE score~\cite{lalor2025simtoreal} as our evaluation metric, calculated via Eq.~\ref{eq:APE_score}. The APE score provides a simple, bounded $[0,1]$ agreement metric (0=maximal deviation, 1=perfect reconstruction) to describe how precisely a model reconstructs the mixed isotopic composition of a heterogeneous source.

\begin{equation}
\mathrm{APE}(h_{\theta,\phi};\mathcal{D}_{\mathrm{test}})
\coloneqq
1-\frac{1}{2n_{\mathrm{test}}}\sum_{i=1}^{n_{\mathrm{test}}}
\left\| \hat y_i - y_i \right\|_1~,
\qquad
\hat y_i \coloneqq h_{\theta,\phi}(x_i)\in\Delta^K,\ \ y_i\in\Delta^K~.
\label{eq:APE_score}
\end{equation}

We also include a ``Train on Target'' row in Table~\ref{table:UDA_scores}, indicating the performance of a model trained directly on a labeled target-domain dataset, serving as a theoretical upper bound on model performance and included only for reference. We remark that Train on Target = 1 for the sim-to-real scenarios, stemming from the single-label, high SNR nature of the dataset. While a more complicated experimental dataset is desirable, we still expect the results of this research to apply broadly and suggest analysis using more challenging datasets as an avenue for future work.

We observe mixed results in the sim-to-sim scenario. The most consistent performance gain compared to source-only were the DANN models, where the MLP achieved an APE score of $0.649 \pm 0.004$ (compared to $0.623 \pm 0.002$ of source-only), TBNN (Li \emph{et al.}) achieved an APE score of $0.690 \pm 0.005$ (compared to $0.630 \pm 0.003$ of source-only), TBNN-LinEmb achieved an APE score of $0.691 \pm 0.006$ (compared to $0.653 \pm 0.004$ of source-only), and TBNN-NonlinEmb achieved an APE score of $0.697 \pm 0.006$ (compared to $0.668 \pm 0.005$ of source-only). DAN and ADDA yielded similar improvements, although the performance gain was only seen using the three transformer architectures. None of the UDA methods improved performance using a CNN backbone.

In the sim-to-real dataset, we observe a stronger result. Across all architectural backbones, the best UDA model achieves an average classification accuracy improvement of $14.9$ percentage points relative to the source-only model. DANN and DAN yielded the strongest average improvement except when using a CNN backbone, in which case DeepCORAL was best. Using the \LaBr{} testing set, the DAN MLP achieved an accuracy of $0.895 \pm 0.018$ (versus $0.743 \pm 0.010$ source-only), DeepCORAL CNN achieved $0.937 \pm 0.021$ (versus $0.800 \pm 0.019$ source-only), DANN TBNN (Li \emph{et al.}) achieved $0.932 \pm 0.022$ (versus $0.775 \pm 0.010$ source-only), DAN TBNN-LinEmb achieved $0.904 \pm 0.022$ (versus $0.754 \pm 0.014$ source-only), and DANN TBNN-NonlinEmb achieved $0.934 \pm 0.022$ (versus $0.765 \pm 0.013$ baseline).

\begin{table}
\centering
\caption{Evaluation metric scores on the target-domain test dataset for seven different UDA techniques. The ``Source-only'' row indicates models that were only trained on source-domain data with no domain adaptation step, serving as a baseline. The ``Train on Target'' row indicates models that were trained on labeled target-domain data, serving as a theoretical best-case scenario. Cell entries are calculated as the mean and sample standard deviation across 10 random trials. The best performing method for each architecture (excluding ``Train on Target'') is highlighted in bold.}
\label{table:UDA_scores}
\begin{subtable}{\textwidth}
\centering
\caption{Scenario: sim-to-sim (\HPGe). Cell entries indicate testing APE score}
\label{table:UDA_scores_hpge}
\begin{tabular}{l c c c c c}
\toprule
 & MLP & CNN & TBNN (Li \emph{et al.}) & TBNN-LinEmb & TBNN-NonlinEmb \\
\midrule
Source-only  & 0.623 $\pm$ 0.002 & \textbf{0.635} $\pm$ \textbf{0.001} & 0.630 $\pm$ 0.003 & 0.653 $\pm$ 0.004 & 0.668 $\pm$ 0.005 \\
ADDA         & 0.604 $\pm$ 0.012 & 0.614 $\pm$ 0.018 & 0.667 $\pm$ 0.007 & 0.678 $\pm$ 0.009 & 0.688 $\pm$ 0.008 \\
DAN          & 0.592 $\pm$ 0.004 & 0.607 $\pm$ 0.003 & 0.659 $\pm$ 0.005 & 0.667 $\pm$ 0.004 & 0.691 $\pm$ 0.005 \\
DANN         & \textbf{0.649} $\pm$ \textbf{0.004} & 0.630 $\pm$ 0.012 & \textbf{0.690} $\pm$ \textbf{0.005} & \textbf{0.691} $\pm$ \textbf{0.006} & \textbf{0.697} $\pm$ \textbf{0.006} \\
DeepCORAL    & 0.629 $\pm$ 0.002 & 0.619 $\pm$ 0.012 & 0.641 $\pm$ 0.012 & 0.640 $\pm$ 0.009 & 0.650 $\pm$ 0.006 \\
DeepJDOT     & 0.390 $\pm$ 0.185 & 0.614 $\pm$ 0.005 & 0.631 $\pm$ 0.005 & 0.626 $\pm$ 0.004 & 0.609 $\pm$ 0.011 \\
Mean Teacher  & 0.484 $\pm$ 0.014 & 0.573 $\pm$ 0.009 & 0.589 $\pm$ 0.112 & 0.633 $\pm$ 0.016 & 0.612 $\pm$ 0.026 \\
SimCLR       & 0.573 $\pm$ 0.005 & 0.605 $\pm$ 0.007 & 0.649 $\pm$ 0.004 & 0.605 $\pm$ 0.006 & 0.648 $\pm$ 0.009 \\
Train on Target & 0.896 $\pm$ 0.001 & 0.881 $\pm$ 0.001 & 0.891 $\pm$ 0.001 & 0.895 $\pm$ 0.001 & 0.889 $\pm$ 0.001 \\
\bottomrule
\end{tabular}
\end{subtable}

\vspace{1.5em}
\begin{subtable}{\textwidth}
\centering
\caption{Scenario: sim-to-real (\LaBr). Cell entries indicate testing accuracy}
\label{table:UDA_scores_labr}
\begin{tabular}{l c c c c c}
\toprule
 & MLP & CNN & TBNN (Li \emph{et al.}) & TBNN-LinEmb & TBNN-NonlinEmb \\
\midrule
Source-only & 0.743 $\pm$ 0.010 & 0.800 $\pm$ 0.019 & 0.775 $\pm$ 0.010 & 0.754 $\pm$ 0.014 & 0.765 $\pm$ 0.013 \\
ADDA & 0.807 $\pm$ 0.039 & 0.868 $\pm$ 0.014 & 0.836 $\pm$ 0.036 & 0.821 $\pm$ 0.044 & 0.826 $\pm$ 0.036 \\
DAN & \textbf{0.895 $\pm$ 0.018} & 0.841 $\pm$ 0.028 & 0.901 $\pm$ 0.027 & \textbf{0.904 $\pm$ 0.022} & 0.878 $\pm$ 0.019 \\
DANN & 0.879 $\pm$ 0.029 & 0.926 $\pm$ 0.016 & \textbf{0.932} $\pm$ \textbf{0.022} & 0.898 $\pm$ 0.032 & \textbf{0.934} $\pm$ \textbf{0.022} \\
DeepCORAL & 0.796 $\pm$ 0.028 & \textbf{0.937 $\pm$ 0.021} & 0.853 $\pm$ 0.017 & 0.876 $\pm$ 0.026 & 0.854 $\pm$ 0.024 \\
DeepJDOT & 0.747 $\pm$ 0.015 & 0.818 $\pm$ 0.020 & 0.800 $\pm$ 0.032 & 0.830 $\pm$ 0.038 & 0.849 $\pm$ 0.030 \\
Mean Teacher & 0.757 $\pm$ 0.010 & 0.773 $\pm$ 0.032 & 0.866 $\pm$ 0.026 & 0.746 $\pm$ 0.013 & 0.785 $\pm$ 0.021 \\
SimCLR & 0.837 $\pm$ 0.014 & 0.869 $\pm$ 0.010 & 0.868 $\pm$ 0.040 & 0.840 $\pm$ 0.032 & 0.853 $\pm$ 0.030 \\
Train on Target & 1.000 $\pm$ 0.000 & 1.000 $\pm$ 0.000 & 1.000 $\pm$ 0.000 & 1.000 $\pm$ 0.000 & 1.000 $\pm$ 0.000 \\
\bottomrule
\end{tabular}
\end{subtable}

\vspace{1.5em}
\begin{subtable}{\textwidth}
\centering
\caption{Scenario: sim-to-real (\NaI). Cell entries indicate testing accuracy}
\label{table:UDA_scores_nai}
\begin{tabular}{l c c c c c}
\toprule
 & MLP & CNN & TBNN (Li \emph{et al.}) & TBNN-LinEmb & TBNN-NonlinEmb \\
\midrule
Source-only & 0.748 $\pm$ 0.011 & 0.818 $\pm$ 0.027 & 0.761 $\pm$ 0.012 & 0.749 $\pm$ 0.018 & 0.742 $\pm$ 0.015 \\
ADDA & 0.841 $\pm$ 0.034 & 0.847 $\pm$ 0.050 & 0.851 $\pm$ 0.030 & 0.828 $\pm$ 0.035 & 0.739 $\pm$ 0.050 \\
DAN & 0.853 $\pm$ 0.028 & 0.833 $\pm$ 0.034 & \textbf{0.933 $\pm$ 0.025} & \textbf{0.905 $\pm$ 0.023} & 0.910 $\pm$ 0.029 \\
DANN & \textbf{0.888} $\pm$ \textbf{0.027} & 0.853 $\pm$ 0.037 & 0.924 $\pm$ 0.022 & 0.898 $\pm$ 0.037 & \textbf{0.921} $\pm$ \textbf{0.023} \\
DeepCORAL & 0.797 $\pm$ 0.031 & \textbf{0.895 $\pm$ 0.025} & 0.877 $\pm$ 0.027 & 0.881 $\pm$ 0.027 & 0.870 $\pm$ 0.029 \\
DeepJDOT & 0.802 $\pm$ 0.016 & 0.798 $\pm$ 0.035 & 0.852 $\pm$ 0.030 & 0.875 $\pm$ 0.026 & 0.882 $\pm$ 0.042 \\
Mean Teacher & 0.739 $\pm$ 0.015 & 0.759 $\pm$ 0.032 & 0.906 $\pm$ 0.018 & 0.776 $\pm$ 0.034 & 0.785 $\pm$ 0.023 \\
SimCLR & 0.841 $\pm$ 0.019 & 0.861 $\pm$ 0.026 & 0.930 $\pm$ 0.016 & 0.849 $\pm$ 0.037 & 0.832 $\pm$ 0.041 \\
Train on Target & 1.000 $\pm$ 0.000 & 1.000 $\pm$ 0.000 & 1.000 $\pm$ 0.000 & 1.000 $\pm$ 0.000 & 1.000 $\pm$ 0.000 \\
\bottomrule
\end{tabular}
\end{subtable}
\end{table}

To statistically support these results, we performed a series of one-sided Wilcoxon signed-rank tests testing the null hypothesis that UDA provides no significant improvement to testing scores compared to the source-only models. We chose a Wilcoxon signed-rank test due to the nature of having paired data (each source-only model was used as the initial configuration for the UDA run) and because the test makes no assumptions regarding the normality of differences. We present the results in Table~\ref{table:UDA_pvalues} for all three domain adaptation scenarios. In the sim-to-sim scenario, the domain-adapted models achieve statistically superior performance to the source-only models in 13 out of 35 comparisons, while in the sim-to-real scenarios, the domain-adapted models achieve statistically superior performance to the source-only models in 30 out of 35 comparisons for the \LaBr{} detector and 27 out of 35 for the \NaI{} detector. In particular, using a transformer model (TBNN (Li \emph{et al.}), TBNN-LinEmb, or TBNN-NonlinEmb) with either DAN or DANN for domain adaptation provides a statistically significant improvement over source-only for every comparison across all domain adaptation scenarios.

\subsection{Latent Representation Analysis}
\label{latent representation analysis}

\begin{figure}
    \centering

    \begin{subfigure}[t]{0.49\textwidth}
        \centering
        \includegraphics[width=\textwidth]{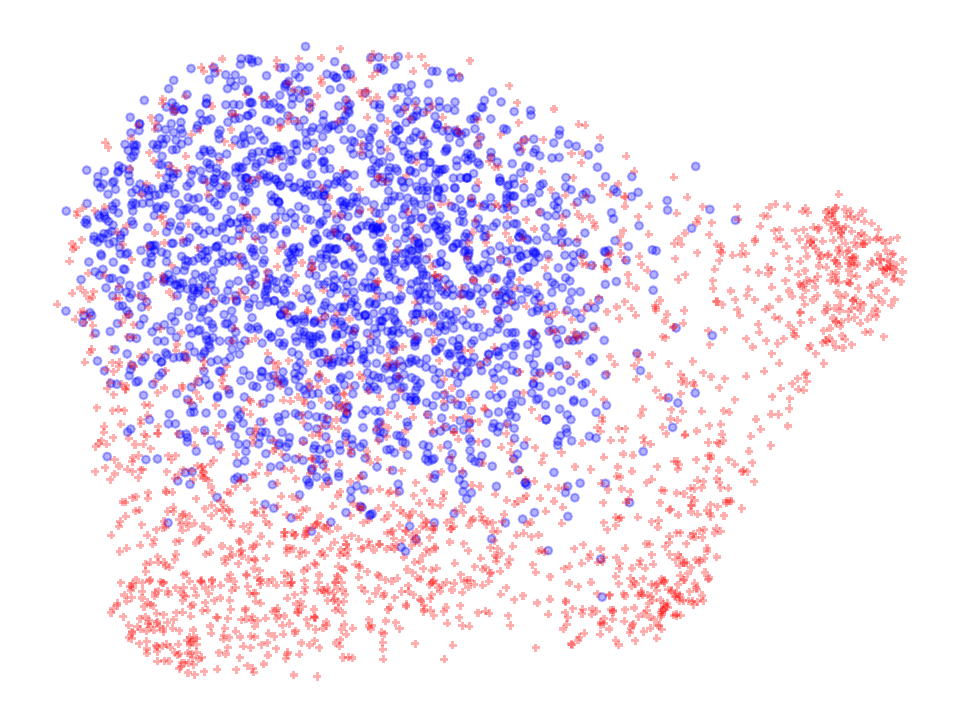}
        \caption{Sim-to-sim \HPGe{} scenario, source-only model.}
        \label{fig:umap_hpge_source}
    \end{subfigure}
    \hfill
    \begin{subfigure}[t]{0.49\textwidth}
        \centering
        \includegraphics[width=\textwidth]{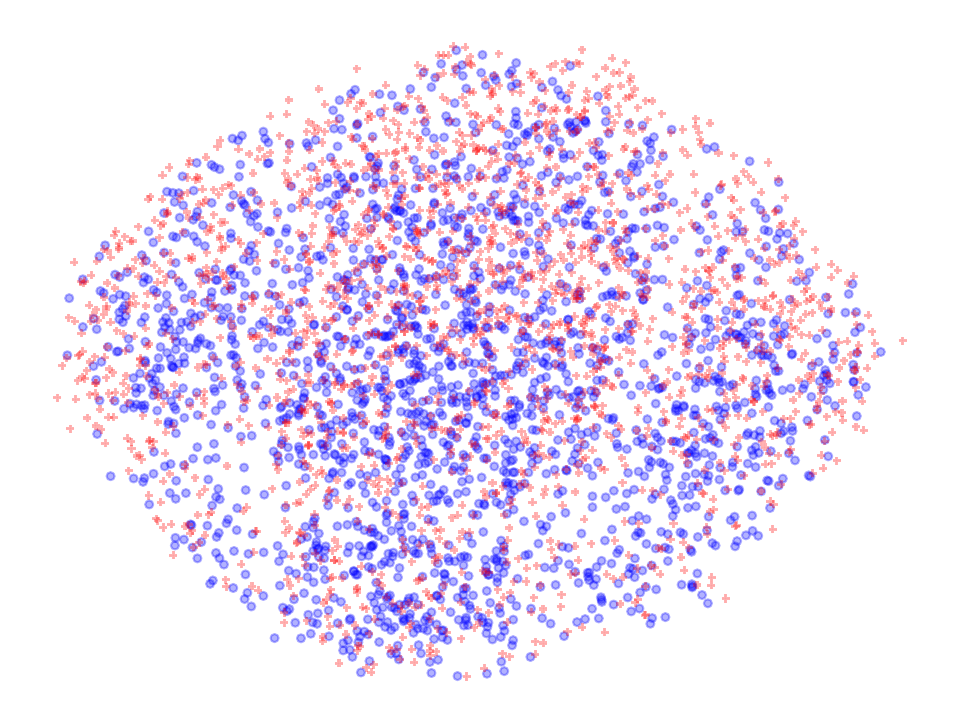}
        \caption{Sim-to-sim \HPGe{} scenario, DANN model.}
        \label{fig:umap_hpge_uda}
    \end{subfigure}

    \vspace{0.5em}

    \begin{subfigure}[t]{0.49\textwidth}
        \centering
        \includegraphics[width=\textwidth]{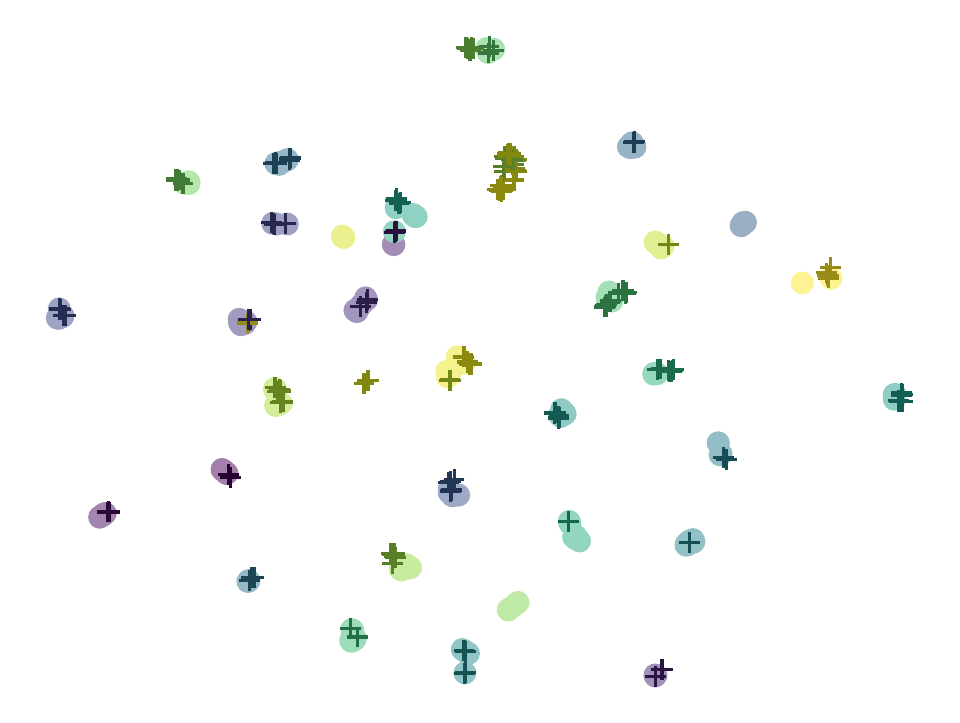}
        \caption{Sim-to-real \LaBr{} scenario, source-only model.}
        \label{fig:umap_labr_source}
    \end{subfigure}
    \hfill
    \begin{subfigure}[t]{0.49\textwidth}
        \centering
        \includegraphics[width=\textwidth]{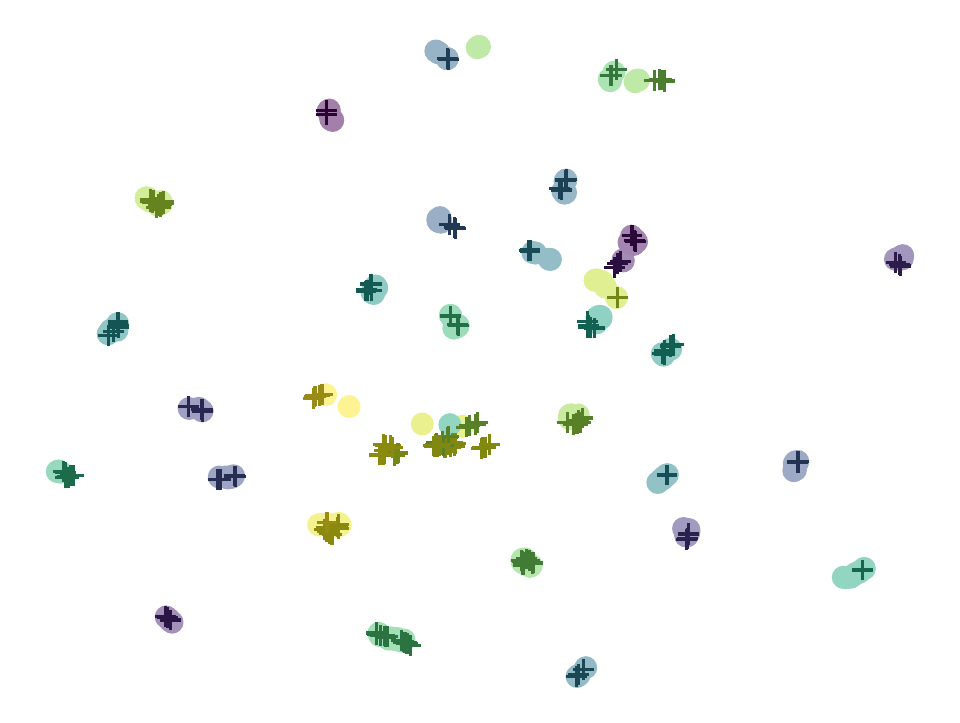}
        \caption{Sim-to-real \LaBr{} scenario, DANN model.}
        \label{fig:umap_labr_uda}
    \end{subfigure}

    \begin{subfigure}[t]{0.49\textwidth}
        \centering
        \includegraphics[width=\textwidth]{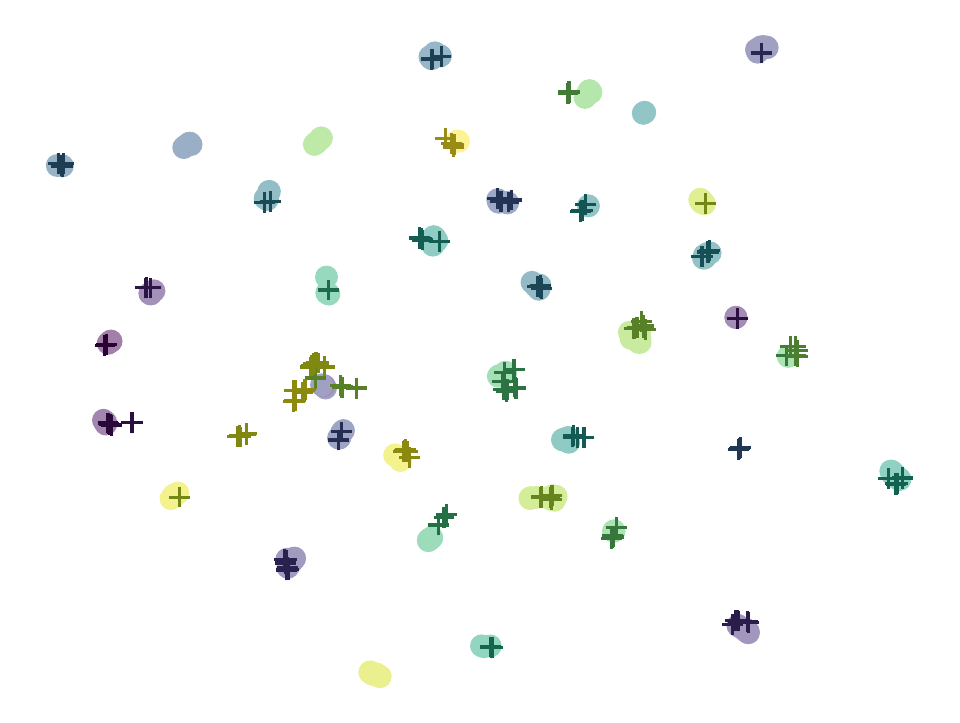}
        \caption{Sim-to-real \NaI{} scenario, source-only model.}
        \label{fig:umap_nai_source}
    \end{subfigure}
    \hfill
    \begin{subfigure}[t]{0.49\textwidth}
        \centering
        \includegraphics[width=\textwidth]{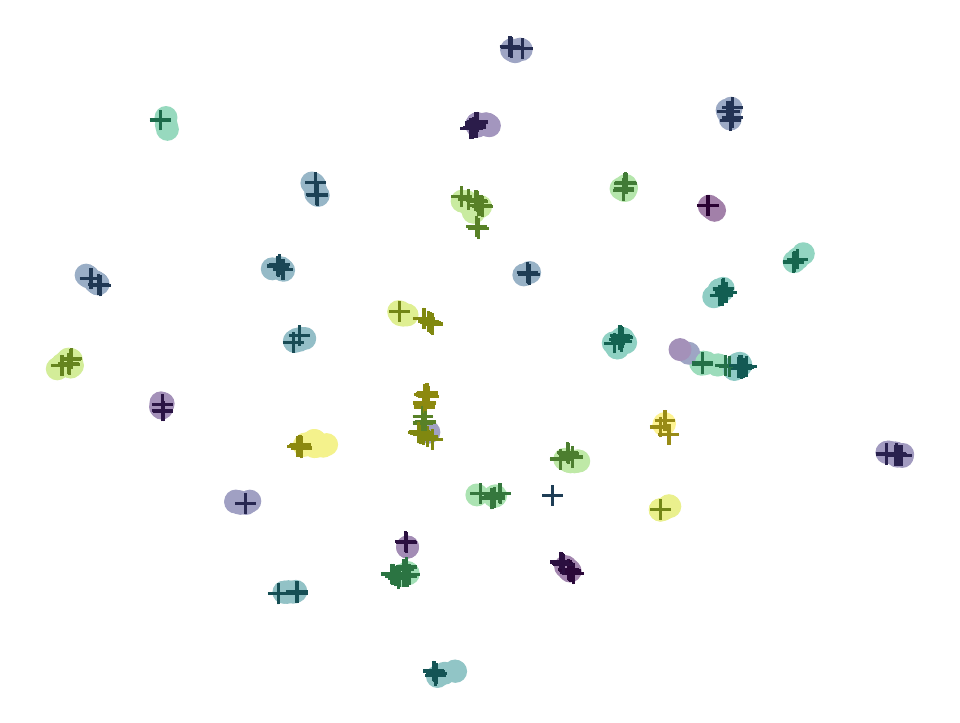}
        \caption{Sim-to-real \NaI{} scenario, DANN model.}
        \label{fig:umap_nai_uda}
    \end{subfigure}

    \caption{UMAP visualizations of the feature extractor outputs for source spectra (`\texttt{o}' markers) and target spectra (`\texttt{+}' markers) using a TBNN-LinEmb architecture. In the sim-to-sim scenario (top row), color indicates domain (blue = source, red = target). In the sim-to-real scenarios (middle and bottom rows), color indicates class, which is possible because the sim-to-real data is single-label.}
    \label{fig:umap_combined}
\end{figure}

To interpret the impact of UDA on model behavior, we visualized the extracted feature representations of source- and target-domain spectra using uniform manifold approximation and projection (UMAP)~\cite{mcinnes2020umapuniformmanifoldapproximation}. We show the results using a TBNN-LinEmb architecture for the sim-to-sim scenario in Figs.~\ref{fig:umap_hpge_source} (source) and~\ref{fig:umap_hpge_uda} (DANN), where we clearly see that the domain adaptation step results in qualitatively better alignment between the extracted source features and target features. Despite the apparent geometric alignment, our results from Section~\ref{testing scores} suggest that the improvement to testing APE score is relatively minor (source-only APE = $0.653 \pm 0.004$, DANN APE = $0.691 \pm 0.006$). In the sim-to-real scenarios, our observations are almost the opposite: Figs.~\ref{fig:umap_labr_source}--\ref{fig:umap_nai_uda} reveal that UDA does not drastically improve qualitative feature alignment. We see strong clustering of both synthetic and experimental features, and it is difficult to articulate improved cluster alignment after domain adaptation. Despite this, we observe substantial double-digit improvements to testing accuracy (source-only accuracy = $0.754 \pm 0.014$, DANN accuracy = $0.898 \pm 0.032$). For reference, we include the same visualizations for other UDA architectures in Fig.~\ref{fig:umap_hpge_all} (sim-to-sim \HPGe{}), Fig.~\ref{fig:umap_labr_all} (sim-to-real \LaBr{}), and Fig.~\ref{fig:umap_nai_all} (sim-to-real \NaI{}).

\subsection{Model Diagnostics}
\label{model diagnostics}

To better understand the performance of our models, we consider a wide range of literature-grounded diagnostic measurements. \texttt{Accuracy} measures the fraction of correctly classified testing spectra, and \texttt{Negative log-likelihood}~\cite{murphy2012machine} reports the average negative log-probability assigned to the true class. \texttt{Brier score}~\cite{glenn1950verification} measures the mean squared error between predicted probabilities and one-hot labels, and \texttt{Expected calibration error}~\cite{guo2017calibration} calculates the average absolute difference between accuracy and confidence across probability bins. \texttt{Prediction margin mean (p10)}~\cite{bartlett1998boosting} measures the mean ($10^\text{th}$) percentile of the log-probability margin between true and best alternative class, and \texttt{AUROC (entropy)} evaluates the ability of predictive uncertainty to identify misclassified examples using the area under the receiver operating characteristic of the uncertainty score~\cite{hendrycks2018baselinedetectingmisclassifiedoutofdistribution}. \texttt{Mean Jacobian norm} measures input sensitivity by calculating the mean squared $\ell_2$ norm of the loss gradient with respect to the normalized spectral inputs~\cite{goodfellow2014explaining}. Lastly, we construct a $k$-nearest-neighbor ($k$-NN) graph of the input spectra and compute the following metrics:~\cite{belkin2006manifold,zhu2003semi} (i) \texttt{$k$-NN TV Hard} is the fraction of $k$-NN edges with differing predicted class labels, (ii) \texttt{$k$-NN prob $\ell_2$} is the mean squared $\ell_2$ distance between prediction probability vectors of $k$-NN neighbors, (iii) \texttt{$k$-NN conf absdiff} is the mean absolute difference in top-1 confidence between $k$-NN neighbors, and (iv) \texttt{$k$-NN margin absdiff} is the mean absolute difference in true-class log-probability margin between $k$-NN neighbors.

We show the results in Table~\ref{table:metric-comparison} for a TBNN-LinEmb model for all three domain adaptation scenarios using the trained source-only and DAN models. We omit hard-label metrics from the sim-to-sim analysis because the dataset uses multi-proportion (soft) labels. Uncertainties ($\pm$) represent sample standard deviations across the 10 randomized trials described in Section~\ref{model training}. We tested for statistical significance using a two-sided Wilcoxon signed-rank test, testing the null hypothesis that there is no difference in diagnostic metric value between the two models. For the sim-to-real \LaBr{} scenario, we find that the DAN model wins in 9 out of 12 metrics, the source-only model in 1 out of 12, and 2 out of 12 considered statistically insignificant. We see similar results for the sim-to-real \NaI{} scenario (DAN wins 7, source-only wins 1, 4 statistically insignificant). Overall, we present statistical evidence that in the sim-to-real scenarios, UDA improves model performance in a range of calibration, separation, uncertainty, and input-space smoothness metrics.

\begin{table}
\centering
\caption{Comparing the effects of UDA on a variety of diagnostic metrics ($\uparrow$ higher is better, $\downarrow$ lower is better). Source-only and UDA TBNN-LinEmb models are shown for all three domain adaptation scenarios. Entries for the best model are bolded when significant ($p < 0.01$). Overall, UDA provides improvements using a wide variety of metrics beyond classification accuracy in the sim-to-real scenarios.}
\label{table:metric-comparison}

\begin{subtable}{\textwidth}
\centering
\caption{Scenario: sim-to-sim (\HPGe).}
\label{table:metric-comparison-hpge}
\adjustbox{max width=\linewidth}{
\begin{tabular}{l l l c c l}
\toprule
\textbf{Metric} & \textbf{Source-only} & \textbf{DAN} & \textbf{$p$-value} & \textbf{Dir.} & \textbf{Type} \\
\midrule
\texttt{APE Score} & $0.653 \pm 0.004$ & $\mathbf{0.667 \pm 0.004}$ & $\mathbf{0.002}$ & $\uparrow$ & Task Score \\
\texttt{Negative log-likelihood}~\cite{murphy2012machine} & $3.499 \pm 0.051$ & $\mathbf{3.180 \pm 0.019}$ & $\mathbf{0.002}$ & $\downarrow$ & Likelihood / Loss \\
\texttt{Brier score}~\cite{glenn1950verification} & $0.056 \pm 0.003$ & $0.057 \pm 0.002$ & $0.322$ & $\downarrow$ & Calibration \\
\texttt{Mean Jacobian norm}~\cite{goodfellow2014explaining} & $0.106 \pm 0.010$ & $\mathbf{0.066 \pm 0.008}$ & $\mathbf{0.002}$ & $\downarrow$ & Sensitivity \\
\texttt{$k$-NN prob $\ell_2$}~\cite{belkin2006manifold,zhu2003semi} & $\mathbf{0.200 \pm 0.010}$ & $0.233 \pm 0.006$ & $\mathbf{0.002}$ & $\downarrow$ & Smoothness \\
\texttt{$k$-NN conf absdiff}~\cite{belkin2006manifold,zhu2003semi} & $\mathbf{0.100 \pm 0.006}$ & $0.120 \pm 0.007$ & $\mathbf{0.002}$ & $\downarrow$ & Smoothness \\
\bottomrule
\end{tabular}
}
\end{subtable}
\vspace{1.5em}

\begin{subtable}{\textwidth}
\centering
\caption{Scenario: sim-to-real (\LaBr)}
\label{table:metric-comparison-labr}
\adjustbox{max width=\linewidth}{
    \begin{tabular}{l l l c c l}
    \toprule
    \textbf{Metric} & \textbf{Source-only} & \textbf{DAN} & \textbf{$p$-value} & \textbf{Dir.} & \textbf{Type} \\
    \midrule
    \texttt{Accuracy} & $0.754 \pm 0.010$ & $\mathbf{0.904 \pm 0.021}$ & $\mathbf{0.002}$ & $\uparrow$ & Accuracy \\
    \texttt{Negative log-likelihood}~\cite{murphy2012machine} & $2.840 \pm 0.317$ & $\mathbf{0.393 \pm 0.125}$ & $\mathbf{0.002}$ & $\downarrow$ & Likelihood / Loss \\
    \texttt{Brier score}~\cite{glenn1950verification} & $0.455 \pm 0.037$ & $\mathbf{0.147 \pm 0.031}$ & $\mathbf{0.002}$ & $\downarrow$ & Calibration \\
    \texttt{Expected calibration error}~\cite{guo2017calibration} & $0.218 \pm 0.025$ & $\mathbf{0.044 \pm 0.014}$ & $\mathbf{0.002}$ & $\downarrow$ & Calibration \\
    \texttt{Prediction margin (mean)}~\cite{bartlett1998boosting} & $6.646 \pm 0.544$ & $\mathbf{8.853 \pm 1.055}$ & $\mathbf{0.002}$ & $\uparrow$ & Separation / Margin \\
    \texttt{Prediction margin (p10)}~\cite{bartlett1998boosting} & $-13.543 \pm 1.765$ & $\mathbf{0.217 \pm 1.116}$ & $\mathbf{0.002}$ & $\uparrow$ & Separation / Margin \\
    \texttt{AUROC (Entropy)}~\cite{shannon1948mathematical} & $0.902 \pm 0.036$ & $\mathbf{0.940 \pm 0.015}$ & $\mathbf{0.004}$ & $\uparrow$ & Uncertainty \\
    \texttt{Mean Jacobian norm}~\cite{goodfellow2014explaining} & $2.151 \pm 0.498$ & $3.348 \pm 2.423$ & $0.322$ & $\downarrow$ & Sensitivity \\
    \texttt{$k$-NN TV hard}~\cite{belkin2006manifold,zhu2003semi} & $0.959 \pm 0.002$ & $0.959 \pm 0.003$ & $0.515$ & $\downarrow$ & Boundary Complexity \\
    \texttt{$k$-NN prob $\ell_2$}~\cite{belkin2006manifold,zhu2003semi} & $1.880 \pm 0.026$ & $\mathbf{1.839 \pm 0.023}$ & $\mathbf{0.004}$ & $\downarrow$ & Smoothness \\
    \texttt{$k$-NN conf absdiff}~\cite{belkin2006manifold,zhu2003semi} & $\mathbf{0.028 \pm 0.017}$ & $0.054 \pm 0.014$ & $\mathbf{0.002}$ & $\downarrow$ & Smoothness \\
    \texttt{$k$-NN margin absdiff}~\cite{belkin2006manifold,zhu2003semi} & $7.666 \pm 0.426$ & $\mathbf{4.890 \pm 0.649}$ & $\mathbf{0.002}$ & $\downarrow$ & Smoothness \\
    \bottomrule
    \end{tabular}
}
\end{subtable}

\vspace{1.5em}
\begin{subtable}{\textwidth}
\centering
\caption{Scenario: sim-to-real (\NaI)}
\label{table:metric-comparison-NaI}
\adjustbox{max width=\linewidth}{
\begin{tabular}{l l l c c l}
\toprule
\textbf{Metric} & \textbf{Source-only} & \textbf{DAN} & \textbf{$p$-value} & \textbf{Dir.} & \textbf{Type} \\
\midrule
\texttt{Accuracy} & $0.749 \pm 0.014$ & $\mathbf{0.905 \pm 0.021}$ & $\mathbf{0.002}$ & $\uparrow$ & Accuracy \\
\texttt{Negative log-likelihood}~\cite{murphy2012machine} & $2.175 \pm 0.228$ & $\mathbf{0.418 \pm 0.120}$ & $\mathbf{0.002}$ & $\downarrow$ & Likelihood / Loss \\
\texttt{Brier score}~\cite{glenn1950verification} & $0.452 \pm 0.023$ & $\mathbf{0.154 \pm 0.037}$ & $\mathbf{0.002}$ & $\downarrow$ & Calibration \\
\texttt{Expected calibration error}~\cite{guo2017calibration} & $0.217 \pm 0.013$ & $\mathbf{0.061 \pm 0.022}$ & $\mathbf{0.002}$ & $\downarrow$ & Calibration \\
\texttt{Prediction margin (mean)}~\cite{bartlett1998boosting} & $7.906 \pm 0.389$ & $\mathbf{10.239 \pm 0.755}$ & $\mathbf{0.002}$ & $\uparrow$ & Separation / Margin \\
\texttt{Prediction margin (p10)}~\cite{bartlett1998boosting} & $-9.823 \pm 1.128$ & $\mathbf{0.286 \pm 1.021}$ & $\mathbf{0.002}$ & $\uparrow$ & Separation / Margin \\
\texttt{AUROC (Entropy)}~\cite{shannon1948mathematical} & $0.921 \pm 0.011$ & $0.923 \pm 0.023$ & $0.846$ & $\uparrow$ & Uncertainty \\
\texttt{Mean Jacobian norm}~\cite{goodfellow2014explaining} & $4.259 \pm 1.011$ & $3.320 \pm 1.567$ & $0.131$ & $\downarrow$ & Sensitivity \\
\texttt{$k$-NN TV hard}~\cite{belkin2006manifold,zhu2003semi} & $\mathbf{0.963 \pm 0.003}$ & $0.972 \pm 0.001$ & $\mathbf{0.002}$ & $\downarrow$ & Boundary Complexity \\
\texttt{$k$-NN prob $\ell_2$}~\cite{belkin2006manifold,zhu2003semi} & $1.880 \pm 0.016$ & $1.876 \pm 0.024$ & $1.000$ & $\downarrow$ & Smoothness \\
\texttt{$k$-NN conf absdiff}~\cite{belkin2006manifold,zhu2003semi} & $0.032 \pm 0.014$ & $0.043 \pm 0.017$ & $0.232$ & $\downarrow$ & Smoothness \\
\texttt{$k$-NN margin absdiff}~\cite{belkin2006manifold,zhu2003semi} & $7.840 \pm 0.600$ & $\mathbf{5.902 \pm 0.560}$ & $\mathbf{0.002}$ & $\downarrow$ & Smoothness \\
\bottomrule
\end{tabular}
}
\end{subtable}
\end{table}

\subsection{Model Explainability}
\label{model explainability}

\begin{figure}
    \centering

    \begin{subfigure}[t]{0.49\textwidth}
        \centering
        \includegraphics[width=\linewidth]{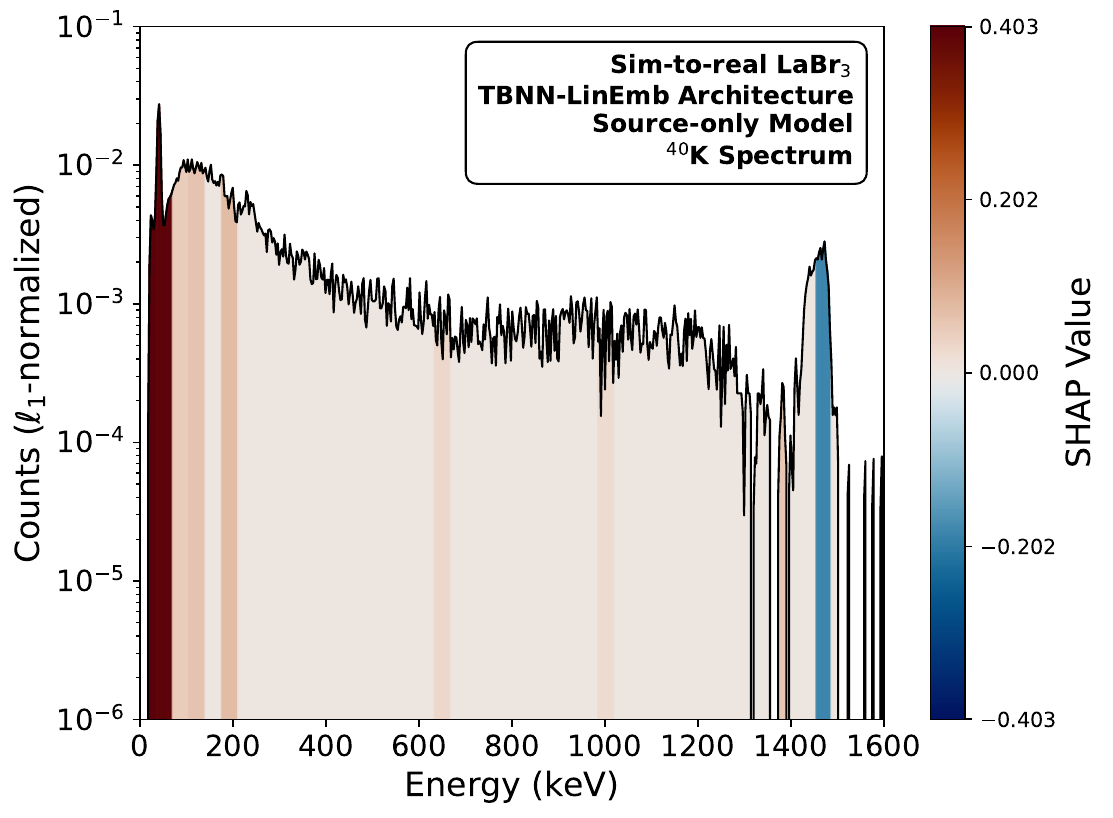}
        \caption{$^{40}$K, source-only (\LaBr{})}
        \label{fig:XAI_K40_src}
    \end{subfigure}
    \hfill
    \begin{subfigure}[t]{0.49\textwidth}
        \centering
        \includegraphics[width=\linewidth]{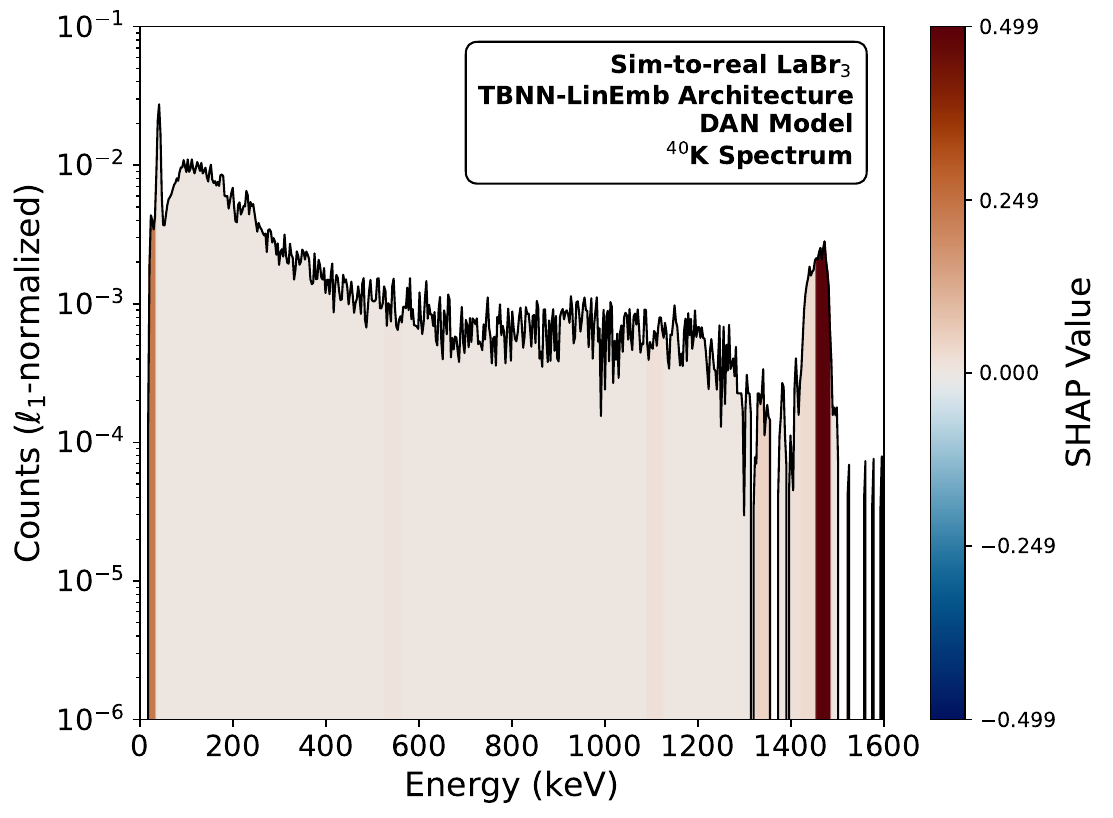}
        \caption{$^{40}$K, DAN (\LaBr{})}
        \label{fig:XAI_K40_uda}
    \end{subfigure}

    \vspace{0.5em}

    \begin{subfigure}[t]{0.49\textwidth}
        \centering
        \includegraphics[width=\linewidth]{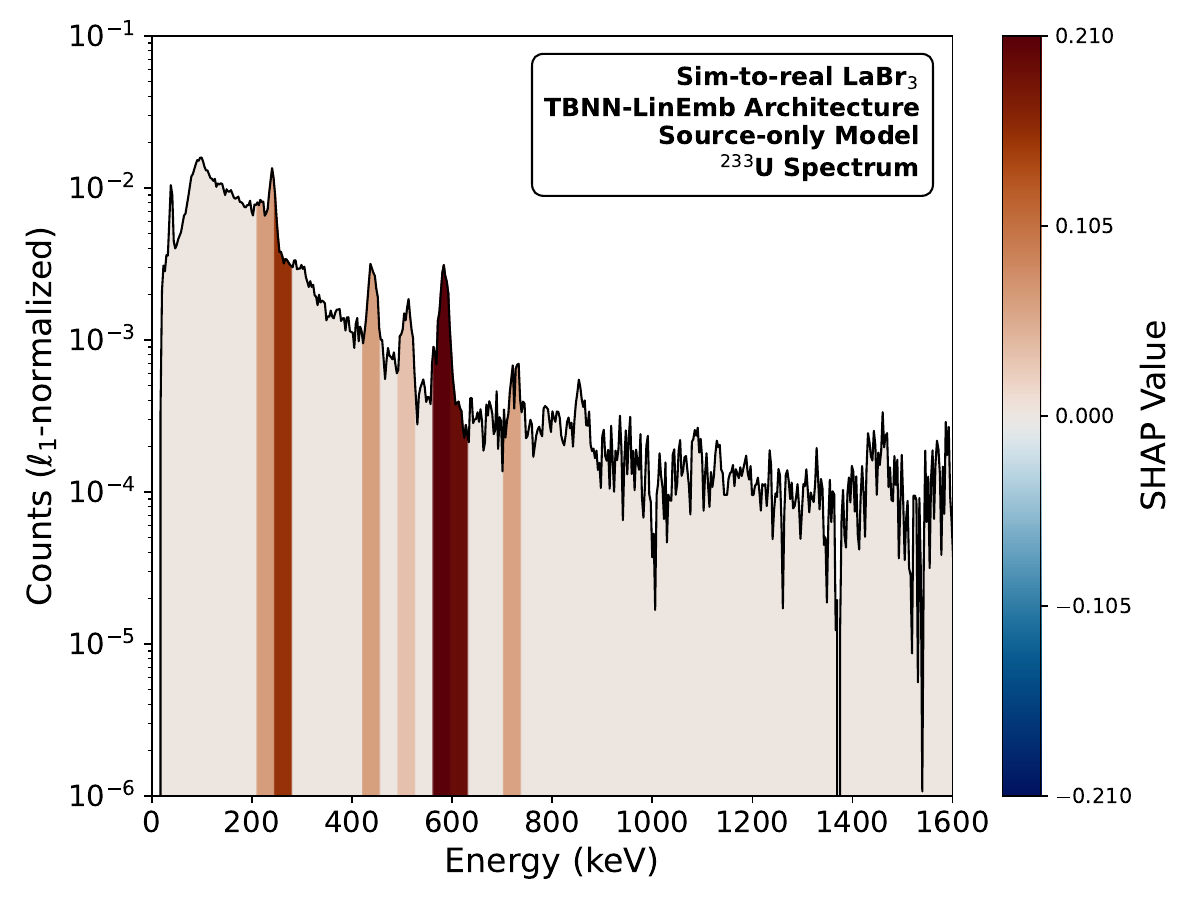}
        \caption{$^{233}$U, source-only (\LaBr{})}
        \label{fig:XAI_U233_src}
    \end{subfigure}
    \hfill
    \begin{subfigure}[t]{0.49\textwidth}
        \centering
        \includegraphics[width=\linewidth]{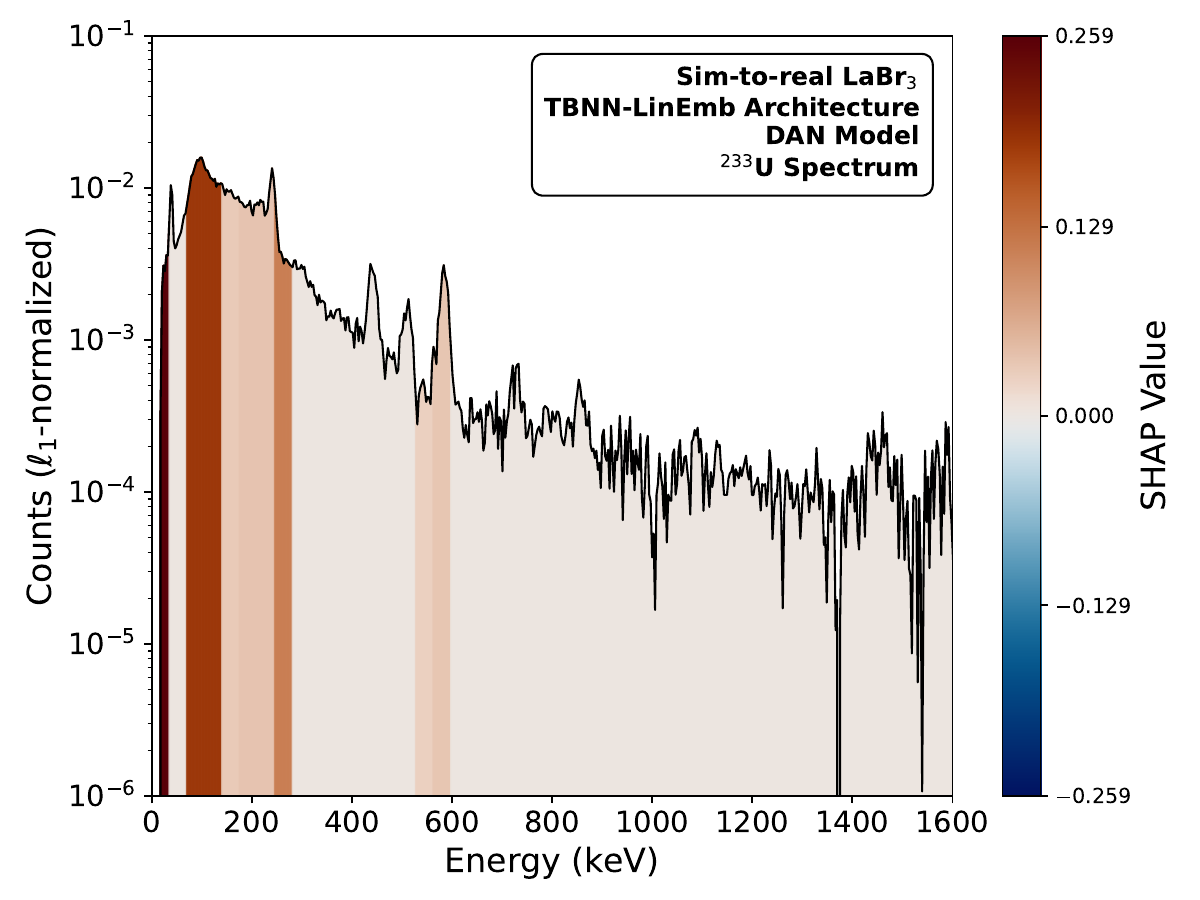}
        \caption{$^{233}$U, DAN (\LaBr{})}
        \label{fig:XAI_U233_uda}
    \end{subfigure}

    \vspace{0.5em}

    \begin{subfigure}[t]{0.49\textwidth}
        \centering
        \includegraphics[width=\linewidth]{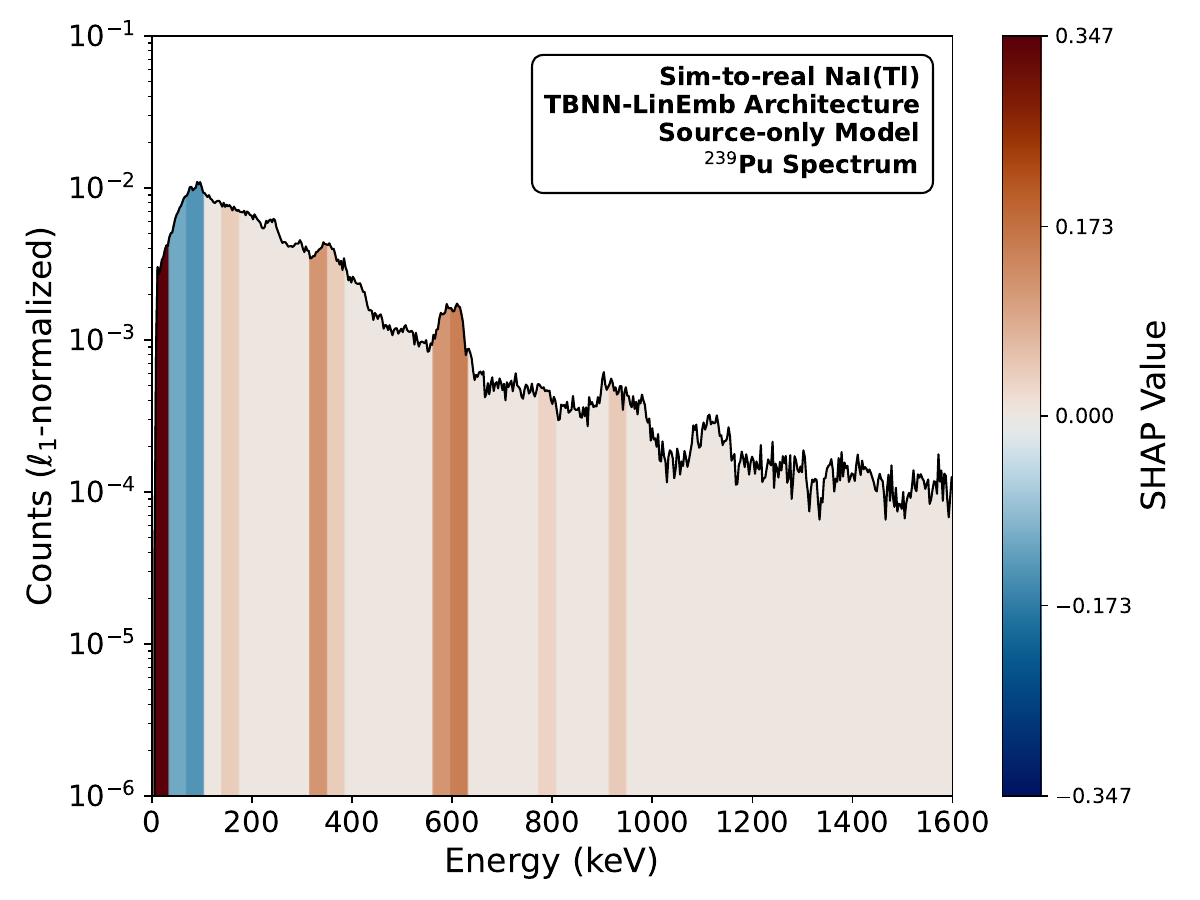}
        \caption{$^{239}$Pu, source-only (\NaI{})}
        \label{fig:XAI_Pu239_src}
    \end{subfigure}
    \hfill
    \begin{subfigure}[t]{0.49\textwidth}
        \centering
        \includegraphics[width=\linewidth]{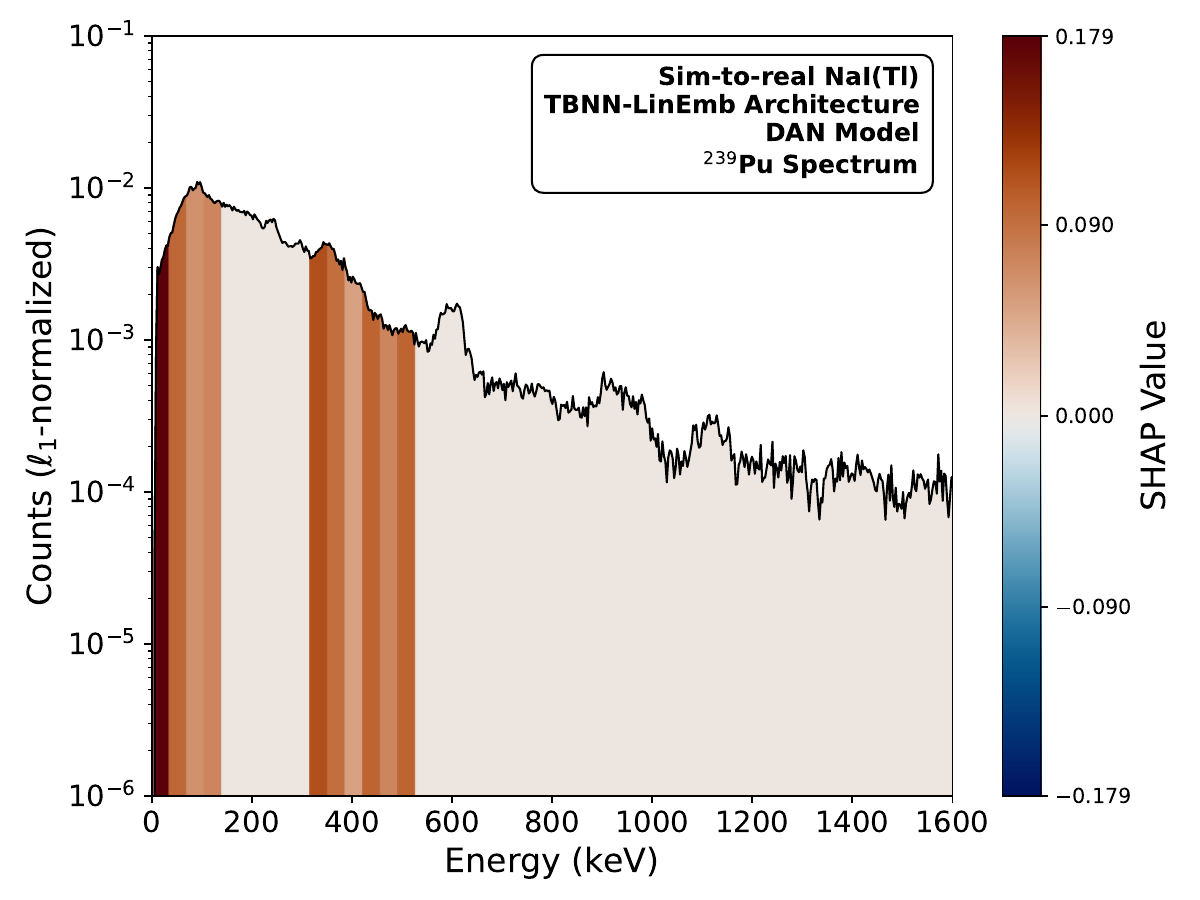}
        \caption{$^{239}$Pu, DAN (\NaI{})}
        \label{fig:XAI_Pu239_uda}
    \end{subfigure}

    \caption{SHAP explanations comparing source-only (left panels) and DAN models (right panels) across three experimental example spectra. For $^{40}$K (top row), the source-only model highlights the \LaBr{} detector's intrinsic 32 keV X-ray peak, whereas the DAN model instead identifies the correct 1460 keV line. For $^{233}$U (middle row), the source-only model relies primarily on peaks associated with visible $^{232}$U-chain contamination, while the DAN model identifies the more subtle 93 keV feature of $^{233}$U. For $^{239}$Pu (bottom row), the source-only model predicts $^{124}$Sb by emphasizing the 603 keV peak (likely due to trace contamination or activation-related background), whereas the DAN model highlights the characteristic $^{239}$Pu peaks around 345, 375, and 413 keV.}
    \label{fig:XAI}
    
\end{figure}

To gain insight into our models' decision-making process, we used SHapley Additive exPlanations (SHAP)~\cite{bandstra2023explaining, lundberg2017unifiedapproachinterpretingmodel} to compute per-region feature attributions. For a given gamma spectrum, SHAP produces an \textit{explanation vector} of SHAP values by assigning an attribution to each spectral region, indicating how that region shifts the model predicted isotope probability relative to a baseline. The SHAP values are computed by repeatedly masking random regions of a spectrum, evaluating the model on the perturbed spectrum, and then averaging each region's marginal contribution. In this context, we mask a region by replacing the counts in the selected bins with values obtained by log-linear interpolation between the region’s endpoints. Regions with large positive (negative) SHAP values increase (decrease) the predicted probability for a given isotope, while values near zero indicate little average effect. We find that, compared to the source-only model, the domain-adapted model is more likely to identify physically descriptive features of the true isotope rather than spurious peaks arising from contamination, background, or detector intrinsics. We include three such examples in Fig.~\ref{fig:XAI}, where we use a colorbar to overlay the computed SHAP values for the model's predicted class on top of the experimental spectrum, comparing the source-only models to the domain-adapted models for different isotopes and detector types.

\section{Discussion}
\label{discussion}

Section~\ref{domain shift characterization} highlights the underlying motivation for this study: simulated and experimental gamma spectra have a quantifiable gap in data distributions. From a physical standpoint, the sim-to-real gap arises from a range of confounding factors: imperfect detector response modeling, unpredictable backgrounds, environmental contributions, complex geometric effects, shielding, class imbalance, and so on. These effects distort spectra in a way that will inevitably degrade the performance of a machine learning model trained on simulations and tested on experimental data. Section~\ref{testing scores} reveals that this domain gap can be partially overcome using UDA. Succinctly, UDA is able to target global feature distribution mismatch without imposing strong assumptions about the nature of the shift. Intuitively, the domain-adapted models have learned how to represent experimental data in a way that matches the patterns learned from simulations.

We remark that the magnitude of the improvement from UDA was substantially different between the sim-to-sim and sim-to-real domain adaptation scenarios. In sim-to-sim, we only observed a modest improvement to testing APE score. This may be due to substantial concept shift, but could also stem from the increased difficulty of 55-class multi-label proportion estimation. The sim-to-real scenario yielded a more significant improvement to testing accuracy, which was initially surprising due to the minimal qualitative improvement to feature alignment as seen in Fig.~\ref{fig:umap_labr_source}--\ref{fig:umap_nai_uda}. We interpret this as evidence that the domain gap in the sim-to-real scenario does not manifest itself as poor 2D UMAP cluster alignment, but rather in the form of complex, high-dimensional feature incongruence which is relevant to the decision boundary.

Furthermore, we observe a performance discrepancy between different UDA methods. DAN and DANN appear to provide the most consistent improvements across a wide range of architectures, likely stemming from their simplicity and minimal assumptions regarding domain shift type. In contrast, augmentation-based methods such as Mean Teacher and SimCLR achieve less consistent gains, suggesting that Poisson resampling alone is insufficient to capture systematic distortions which contribute to the domain gap. This hypothesis is strengthened by the lack of apparent domain alignment shown in Figs.~\ref{fig:umap_hpge_MeanTeacher} and~\ref{fig:umap_hpge_SimCLR}. Interestingly, DeepCORAL's impressive results were limited to the CNN backbone, suggesting covariance-based alignment effectively captures the hierarchical, spatially-structured features with strong local correlations produced by CNNs.

Finally, SHAP (Section~\ref{model explainability}) helps us understand and explain the decision making process of our models, providing further intuition behind the improvements from UDA. For instance, Fig.~\ref{fig:XAI_K40_src} reveals that the source-only model identifies the \LaBr{} detector's intrinsic 32 keV Ba K-shell X-ray peak as salient, confusing it with the 40 keV X-rays emitted from electron capture on $^{152}$Eu. On the contrary, Fig.~\ref{fig:XAI_K40_uda} demonstrates that the DAN model correctly learns to ignore the 32 keV characteristic X-ray and instead focus on the 1460 keV peak from $^{40}$K. As a result, the DAN model achieves a $^{40}$K classification accuracy of 83\%, compared to 17\% from the source-only model. We make similar observations in Figs.~\ref{fig:XAI_U233_src} and~\ref{fig:XAI_Pu239_src}, where the source-only model focuses on miscellaneous peaks stemming from trace contamination, likely because these contaminants were not included in the source-domain simulations and thus the model did not learn to ignore them. On the contrary, we see in Figs.~\ref{fig:XAI_U233_uda} and~\ref{fig:XAI_Pu239_uda} that the domain-adapted model has learned to instead focus on features that are stable across simulation and experiment.

\section{Conclusions}

This study uses UDA to improve the performance of radioisotope identification models in scenarios where target-domain data is available but lacks ground-truth labels. We compared several different UDA methods (ADDA, DAN, DANN, DeepCORAL, DeepJDOT, Mean Teacher, SimCLR) and architectural backbones (MLP, CNN, TBNN), finding that UDA provides a statistically significant improvement to testing scores in both sim-to-sim and sim-to-real scenarios. These improvements were more pronounced in the sim-to-real studies, where we found that DAN and DANN reliably yielded double digit improvements to testing accuracy across the MLP and TBNN model families. For instance, the DAN TBNN-LinEmb achieves a testing accuracy of $0.904 \pm 0.022$ on the experimental \LaBr{} dataset, compared to $0.754 \pm 0.014$ of the source-only model. The improvements in the sim-to-sim scenario were less pronounced but still statistically significant, validating the hypothesis that domain shift in sim-to-sim was more challenging to resolve using unsupervised feature alignment.

We support the conclusions of this study using a range of quantitative diagnostic metrics and qualitative explainability techniques. In addition to testing accuracy, DAN achieves improved likelihood, calibration, and margin compared to the source-only model on the sim-to-real \LaBr{} dataset. Furthermore, we have identified cases where the DAN model highlights salient spectral features instead of overfitting to spurious high-count, lower-energy peaks. Overall, we find that UDA is a practical tool for adapting synthetic-pretrained ML methods for radioisotope identification tasks in operational scenarios where unlabeled experimental datasets are available.

We identify several avenues for future work. First, it would be valuable to consider more challenging sim-to-real scenarios, including mixed sources and low-SNR datasets. Second, follow-on investigations should analyze the effect of dataset size on UDA effectiveness. Third, a more comprehensive SHAP analysis across all isotopes and shielding configurations, potentially summarized using a global SHAP statistic, would be desirable. Fourth, future research should consider additional sources of domain shift, including scenarios where certain isotope classes are missing from the target-domain dataset. Lastly, future work should investigate a \emph{real-to-real} domain adaptation scenario in which data collected from one experiment is used in conjunction with a different but related task.

\section{Data Availability}

Datasets and model training code can be found at \url{https://github.com/pnnl/gamma-adapt/} for the sim-to-sim domain adaptation scenario. Due to data sensitivity, the code, data, and best models for the sim-to-real dataset are not available for public release.

\section{Acknowledgements}

This research was supported by the Laboratory Directed Research and Development Program at Pacific Northwest National Laboratory, a multiprogram national laboratory operated by Battelle for the U.S. Department of Energy under contract DE-AC05-76RLO1830. Peter Lalor is grateful for the support of the Linus Pauling Distinguished Postdoctoral Fellowship. The authors would like to acknowledge Tyler Morrow and Brian Archambault for their useful suggestions and feedback. The authors declare no conflict of interest.

\section{Declaration of generative AI and AI-assisted technologies in the manuscript preparation process}

During the preparation of this work the authors used generative AI tools in order to assist with language editing and formatting. After using this tool/service, the authors reviewed and edited the content as needed and take full responsibility for the content of the published article.

\newpage

\setlength{\bibsep}{0pt}
\bibliography{References.bib}

\newpage

\appendix

\section{Supplementary Materials}
\label{appendix}

\subsection{Hyperparameter optimization}

Throughout the following tables, search spaces are annotated as follows: $[a,\,b]$ denotes continuous uniform sampling; $[a,\,b]_{\log}$ denotes continuous log-uniform sampling; $\{a,\,\ldots,\,b\}$ denotes uniform sampling over consecutive integers; $[a,\,b]_{2^k}$ denotes uniform sampling over powers of two within the interval; and an explicitly listed finite set such as $\{3,5,7,9\}$ indicates uniform sampling over its elements. All hyperparameter searches use the Tree-structured Parzen Estimator (TPE) sampler~\cite{bergstra2011algorithms} implemented in Optuna~\cite{optuna2019}. We observe that some of the optimized hyperparameters in Tables~\ref{table:source_arch_hps}--\ref{table:SimCLR_hps} fall at or near the boundary of their search space. The ranges reported here reflect the final iteration of several rounds of search-space refinement and note that hyperparameters which most strongly affect results (learning rate, UDA tradeoff parameter) are interior to the search space in the vast majority of configurations.

\newpage

\begingroup
\linespread{1.0}\selectfont
\begin{table}
\centering
\caption{Source-only architecture hyperparameter optimization summary for MLP, CNN, TBNN (Li \emph{et al.}), TBNN-LinEmb, and TBNN-NonlinEmb architectures. The table details the search spaces and best-performing hyperparameter values obtained via TPE Bayesian optimization using the source dataset from the sim-to-sim scenario. The best-performing hyperparameters are used for all models across different scenarios in this study. See the search space notation paragraph above for interval and sampling conventions.}
\begin{tabular}{l l l l}
\toprule
Architecture & Parameter           & Search Space      & Best run \\
\midrule
MLP     & Num Dense Layers         & $\{1, \ldots, 4\}$              & 2 \\
        & Dense1 Hidden Units      & $[512,\, 8192]_{2^k}$         & 8192 \\
        & Dense2 Hidden Units      & $[256,\, 4096]_{2^k}$         & 4096 \\
\midrule
CNN     & Num Convolutional Layers & $\{1, 2\}$              & 1 \\
        & Conv Filters             & $[16,\, 256]_{2^k}$           & 16 \\
        & Conv Kernel Size         & $\{3, 5, 7, 9\}$              & 5 \\
        & Num Dense Layers         & $\{1, 2\}$              & 2 \\
        & Dense1 Hidden Units      & $[512,\, 8192]_{2^k}$         & 4096 \\
        & Dense2 Hidden Units      & $[256,\, 4096]_{2^k}$         & 2048 \\
\midrule
TBNN (Li \emph{et al.}) & Num Attention Blocks & $\{1, \ldots, 8\}$   & 4 \\
        & Num Heads                & $[1,\, 8]_{2^k}$              & 4 \\
        & FF Dimension             & $[64,\, 8192]_{2^k}$          & 2048 \\
\midrule
TBNN-LinEmb & Embedding Dimension  & $[8,\, 1024]_{2^k}$           & 256 \\
        & Num Attention Blocks     & $\{1, \ldots, 5\}$              & 5 \\
        & Num Heads                & $[1,\, 8]_{2^k}$              & 2 \\
        & FF Dimension             & $[16,\, 16384]_{2^k}$         & 512 \\
        & Patch Size               & $[16,\, 64]_{2^k}$            & 64 \\
\midrule
TBNN-NonLinEmb & Embedder Hidden Units & $[8,\, 2048]_{2^k}$       & 512 \\
        & Embedding Dimension      & $[8,\, 1024]_{2^k}$           & 512 \\
        & Num Attention Blocks     & $\{1, \ldots, 5\}$              & 3 \\
        & Num Heads                & $[1,\, 8]_{2^k}$              & 4 \\
        & FF Dimension             & $[16,\, 16384]_{2^k}$         & 1024 \\
        & Patch Size               & $[16,\, 64]_{2^k}$            & 64 \\
\bottomrule
\end{tabular}
\label{table:source_arch_hps}
\end{table}
\endgroup

\begingroup
\linespread{1.0}\selectfont
\begin{table}
\centering
\caption{Source-only training hyperparameter optimization summary for MLP, CNN, TBNN (Li \emph{et al.}), TBNN-LinEmb, and TBNN-NonlinEmb architectures. The table details the search spaces and best-performing hyperparameter values obtained via TPE Bayesian optimization using the source dataset. See the search space notation paragraph for interval and sampling conventions.}
\begin{tabular}{l l l l l l}
\toprule
& & & \multicolumn{3}{c}{Best run} \\
\cmidrule(lr){4-6}
 & & & sim-to-sim & sim-to-real & sim-to-real \\
Architecture & Parameter & Search Space & \HPGe & \LaBr & \NaI \\
\midrule
MLP     & Learning rate & $[10^{-5},\, 2{\times}10^{-3}]_{\log}$ & 4.60e-5 & 1.80e-3 & 1.32e-3 \\
        & Batch size    & $[32,\, 512]_{2^k}$    & 256     & 512     & 256 \\
        & Weight decay  & $[10^{-7},\, 10^{-3}]_{\log}$ & 1.54e-5 & 1.96e-6 & 1.21e-4 \\
        & Dropout       & $[0.0,\, 0.4]$   & 0.230   & 0.394   & 0.224 \\
\midrule
CNN     & Learning rate & $[10^{-5},\, 2{\times}10^{-3}]_{\log}$ & 4.17e-5 & 1.07e-5 & 2.94e-4 \\
        & Batch size    & $[32,\, 512]_{2^k}$    & 256     & 512     & 512 \\
        & Weight decay  & $[10^{-7},\, 10^{-3}]_{\log}$ & 5.77e-5 & 6.21e-4 & 1.25e-4 \\
        & Dropout       & $[0.0,\, 0.4]$   & 0.205   & 0.220   & 0.272 \\
\midrule
TBNN (Li \emph{et al.}) & Learning rate & $[10^{-5},\, 2{\times}10^{-3}]_{\log}$ & 1.65e-3 & 1.28e-3 & 2.99e-4 \\
        & Batch size    & $[32,\, 512]_{2^k}$    & 512     & 512     & 256 \\
        & Weight decay  & $[10^{-7},\, 10^{-3}]_{\log}$ & 9.06e-6 & 2.66e-5 & 5.39e-5 \\
        & Dropout       & $[0.0,\, 0.1]$   & 3.31e-5 & 0.0349  & 0.0623 \\
\midrule
TBNN-LinEmb & Learning rate & $[10^{-5},\, 2{\times}10^{-3}]_{\log}$ & 2.75e-4 & 1.01e-4 & 1.54e-4 \\
        & Batch size    & $[32,\, 512]_{2^k}$    & 256     & 512     & 512 \\
        & Weight decay  & $[10^{-7},\, 10^{-3}]_{\log}$ & 1.23e-7 & 2.56e-6 & 8.89e-7 \\
        & Dropout       & $[0.0,\, 0.1]$   & 2.90e-3 & 0.0413  & 0.0737 \\
\midrule
TBNN-NonLinEmb & Learning rate & $[10^{-5},\, 2{\times}10^{-3}]_{\log}$ & 1.94e-4 & 2.27e-5 & 1.32e-4 \\
        & Batch size    & $[32,\, 512]_{2^k}$    & 256     & 256     & 512 \\
        & Weight decay  & $[10^{-7},\, 10^{-3}]_{\log}$ & 2.47e-7 & 6.49e-4 & 3.39e-4 \\
        & Dropout       & $[0.0,\, 0.1]$   & 3.24e-4 & 0.0479  & 0.0947 \\
\bottomrule
\end{tabular}
\label{table:source_training_hps}
\end{table}
\endgroup

\begingroup
\linespread{1.0}\selectfont
\begin{table}
\centering
\caption{ADDA hyperparameter optimization summary for MLP, CNN, TBNN (Li \emph{et al.}), TBNN-LinEmb, and TBNN-NonlinEmb architectures. The table details the search spaces and best-performing hyperparameter values obtained via TPE Bayesian optimization. See the search space notation paragraph for interval and sampling conventions.}
\begin{tabular}{l l l l l l}
\toprule
& & & \multicolumn{3}{c}{Best run} \\
\cmidrule(lr){4-6}
 & & & sim-to-sim & sim-to-real & sim-to-real \\
Architecture & Parameter & Search Space & \HPGe & \LaBr & \NaI \\
\midrule
MLP     & FE learning rate$^{\dagger}$ & $[10^{-8},\, 10^{-2}]_{\log}$      & 5.87e-7 & 1.67e-4 & 2.18e-4 \\
        & DD learning rate$^{\dagger}$ & $[10^{-8},\, 10^{-2}]_{\log}$    & 9.29e-6 & 3.87e-6 & 5.03e-6 \\
        & Batch size                      & $[64,\, 512]_{2^k}$         & 128     & 128     & 64 \\
        & Weight decay                    & $[10^{-7},\, 10^{-3}]_{\log}$      & 6.89e-6 & 2.35e-6 & 8.33e-4 \\
        & Dropout                         & $[0.0,\, 0.4]$        & 0.384   & 0.282   & 0.156 \\
        & Num discriminator layers        & $\{1, 2\}$            & 2       & 2       & 1 \\
        & Discriminator dense1 units      & $[512,\, 4096]_{2^k}$       & 2048    & 4096    & 4096 \\
        & Discriminator dense2 units      & $[256,\, 2048]_{2^k}$       & 1024    & 2048    & N/A \\
\midrule
CNN     & FE learning rate$^{\dagger}$ & $[10^{-8},\, 10^{-2}]_{\log}$      & 4.01e-7 & 6.25e-7 & 3.99e-5 \\
        & DD learning rate$^{\dagger}$ & $[10^{-8},\, 10^{-2}]_{\log}$    & 4.75e-6 & 7.67e-6 & 5.78e-4 \\
        & Batch size                      & $[64,\, 512]_{2^k}$         & 128     & 256     & 256 \\
        & Weight decay                    & $[10^{-7},\, 10^{-3}]_{\log}$      & 6.51e-4 & 4.36e-6 & 4.13e-6 \\
        & Dropout                         & $[0.0,\, 0.4]$        & 0.394   & 0.270   & 0.114 \\
        & Num discriminator layers        & $\{1, 2\}$            & 2       & 2       & 2 \\
        & Discriminator dense1 units      & $[512,\, 4096]_{2^k}$       & 1024    & 512     & 512 \\
        & Discriminator dense2 units      & $[256,\, 2048]_{2^k}$       & 512     & 256     & 256 \\
\midrule
TBNN (Li \emph{et al.})
        & FE learning rate$^{\dagger}$ & $[10^{-8},\, 10^{-2}]_{\log}$      & 3.09e-5 & 5.28e-5 & 2.81e-5 \\
        & DD learning rate$^{\dagger}$ & $[10^{-8},\, 10^{-2}]_{\log}$    & 2.10e-4 & 8.43e-6 & 2.76e-6 \\
        & Batch size                      & $[64,\, 512]_{2^k}$         & 64      & 128     & 512 \\
        & Weight decay                    & $[10^{-7},\, 10^{-3}]_{\log}$      & 2.20e-6 & 9.73e-4 & 4.95e-7 \\
        & Dropout                         & $[0.0,\, 0.4]$        & 6.30e-3 & 0.0561  & 0.103 \\
        & Num discriminator layers        & $\{1, 2\}$            & 1       & 1       & 1 \\
        & Discriminator dense1 units      & $[512,\, 4096]_{2^k}$       & 512     & 1024    & 4096 \\
        & Discriminator dense2 units      & $[256,\, 2048]_{2^k}$       & N/A     & N/A     & N/A \\
\midrule
TBNN-LinEmb
        & FE learning rate$^{\dagger}$ & $[10^{-8},\, 10^{-2}]_{\log}$      & 8.53e-6 & 3.24e-6 & 7.33e-6 \\
        & DD learning rate$^{\dagger}$ & $[10^{-8},\, 10^{-2}]_{\log}$    & 4.21e-4 & 9.38e-6 & 8.65e-6 \\
        & Batch size                      & $[64,\, 512]_{2^k}$         & 256     & 128     & 64 \\
        & Weight decay                    & $[10^{-7},\, 10^{-3}]_{\log}$      & 4.56e-6 & 4.88e-4 & 1.57e-7 \\
        & Dropout                         & $[0.0,\, 0.4]$        & 0.114   & 0.221   & 0.203 \\
        & Num discriminator layers        & $\{1, 2\}$            & 1       & 2       & 2 \\
        & Discriminator dense1 units      & $[512,\, 4096]_{2^k}$       & 4096    & 1024    & 4096 \\
        & Discriminator dense2 units      & $[256,\, 2048]_{2^k}$       & N/A     & 512     & 2048 \\
\midrule
TBNN-NonLinEmb
        & FE learning rate$^{\dagger}$ & $[10^{-8},\, 10^{-2}]_{\log}$      & 1.34e-6 & 5.15e-6 & 4.70e-6 \\
        & DD learning rate$^{\dagger}$ & $[10^{-8},\, 10^{-2}]_{\log}$    & 9.51e-5 & 1.95e-5 & 2.23e-6 \\
        & Batch size                      & $[64,\, 512]_{2^k}$         & 64      & 64      & 512 \\
        & Weight decay                    & $[10^{-7},\, 10^{-3}]_{\log}$      & 4.65e-6 & 2.88e-7 & 1.36e-5 \\
        & Dropout                         & $[0.0,\, 0.4]$        & 8.69e-3 & 0.196   & 0.0285 \\
        & Num discriminator layers        & $\{1, 2\}$            & 2       & 2       & 1 \\
        & Discriminator dense1 units      & $[512,\, 4096]_{2^k}$       & 4096    & 4096    & 2048 \\
        & Discriminator dense2 units      & $[256,\, 2048]_{2^k}$       & 2048    & 2048    & N/A \\
\bottomrule
\end{tabular}
\par\smallskip
{\footnotesize $^{\dagger}$The FE and DD learning rates are jointly parameterized as $\ell_{\mathrm{FE}} = \ell/\sqrt{r}$ and $\ell_{\mathrm{DD}} = \ell\sqrt{r}$, where $\ell \in [10^{-7},\,10^{-3}]_{\log}$ is a shared base learning rate and $r \in [10^{-2},\,10^{2}]_{\log}$ is a ratio parameter. The intervals shown are the effective ranges.}
\label{table:ADDA_hps}
\end{table}
\endgroup

\begingroup
\linespread{1.0}\selectfont
\begin{table}
\centering
\caption{Same as Table~\ref{table:ADDA_hps}, but using DAN.}
\begin{tabular}{l l l l l l}
\toprule
& & & \multicolumn{3}{c}{Best run} \\
\cmidrule(lr){4-6}
 & & & sim-to-sim & sim-to-real & sim-to-real \\
Architecture & Parameter & Search Space & \HPGe & \LaBr & \NaI \\
\midrule
MLP     & Learning rate              & $[10^{-7},\, 10^{-3}]_{\log}$   & 5.80e-6 & 9.72e-4 & 6.84e-4 \\
        & Batch size                 & $[64,\, 512]_{2^k}$      & 256     & 256     & 128     \\
        & Weight decay               & $[10^{-7},\, 10^{-3}]_{\log}$   & 1.92e-5 & 8.15e-5 & 6.24e-6 \\
        & Dropout                    & $[0.0,\, 0.4]$     & 0.0278  & 1.10e-3 & 3.71e-3 \\
        & Tradeoff parameter         & $[10^{-1},\, 10^{4}]_{\log}$    & 7397    & 871     & 1100    \\
        & RBF bandwidth              & $[10^{-1},\, 10^{2}]_{\log}$    & 58.3    & 6.22    & 16.9    \\
        & Number of kernels          & $\{3, 5, 7, \ldots, 15\}$        & 9       & 15      & 11      \\
\midrule
CNN     & Learning rate              & $[10^{-7},\, 10^{-3}]_{\log}$   & 8.73e-4 & 2.17e-4 & 8.04e-4 \\
        & Batch size                 & $[64,\, 512]_{2^k}$      & 64      & 512     & 256     \\
        & Weight decay               & $[10^{-7},\, 10^{-3}]_{\log}$   & 1.85e-4 & 1.42e-6 & 4.27e-7 \\
        & Dropout                    & $[0.0,\, 0.4]$     & 0.283   & 0.396   & 0.381   \\
        & Tradeoff parameter         & $[10^{-1},\, 10^{4}]_{\log}$    & 34.4    & 11.0    & 45.8    \\
        & RBF bandwidth              & $[10^{-1},\, 10^{2}]_{\log}$    & 9.32    & 87.0    & 41.7    \\
        & Number of kernels          & $\{3, 5, 7, \ldots, 15\}$        & 5       & 13      & 13      \\
\midrule
TBNN (Li \emph{et al.})
        & Learning rate              & $[10^{-7},\, 10^{-3}]_{\log}$   & 1.59e-4 & 7.47e-4 & 7.74e-4 \\
        & Batch size                 & $[64,\, 512]_{2^k}$      & 512     & 64      & 256     \\
        & Weight decay               & $[10^{-7},\, 10^{-3}]_{\log}$   & 7.01e-5 & 2.68e-5 & 3.30e-6 \\
        & Dropout                    & $[0.0,\, 0.4]$     & 0.0819  & 0.0198  & 0.213   \\
        & Tradeoff parameter         & $[10^{-1},\, 10^{4}]_{\log}$    & 247     & 286     & 159     \\
        & RBF bandwidth              & $[10^{-1},\, 10^{2}]_{\log}$    & 2.56    & 32.1    & 26.3    \\
        & Number of kernels          & $\{3, 5, 7, \ldots, 15\}$        & 7       & 13      & 15      \\
\midrule
TBNN-LinEmb
        & Learning rate              & $[10^{-7},\, 10^{-3}]_{\log}$   & 2.52e-4 & 3.92e-4 & 4.08e-4 \\
        & Batch size                 & $[64,\, 512]_{2^k}$      & 256     & 512     & 256     \\
        & Weight decay               & $[10^{-7},\, 10^{-3}]_{\log}$   & 7.57e-4 & 5.19e-7 & 1.38e-5 \\
        & Dropout                    & $[0.0,\, 0.4]$     & 0.125   & 0.165   & 0.272   \\
        & Tradeoff parameter         & $[10^{-1},\, 10^{4}]_{\log}$    & 4.31    & 381     & 185     \\
        & RBF bandwidth              & $[10^{-1},\, 10^{2}]_{\log}$    & 3.89    & 1.91    & 0.844   \\
        & Number of kernels          & $\{3, 5, 7, \ldots, 15\}$        & 7       & 9       & 15      \\
\midrule
TBNN-NonLinEmb
        & Learning rate              & $[10^{-7},\, 10^{-3}]_{\log}$   & 5.19e-5 & 4.44e-4 & 4.70e-4 \\
        & Batch size                 & $[64,\, 512]_{2^k}$      & 512     & 512     & 512     \\
        & Weight decay               & $[10^{-7},\, 10^{-3}]_{\log}$   & 3.46e-5 & 9.19e-4 & 7.75e-5 \\
        & Dropout                    & $[0.0,\, 0.4]$     & 0.179   & 0.337   & 0.0733  \\
        & Tradeoff parameter         & $[10^{-1},\, 10^{4}]_{\log}$    & 54.9    & 440     & 168     \\
        & RBF bandwidth              & $[10^{-1},\, 10^{2}]_{\log}$    & 9.58    & 26.9    & 42.1    \\
        & Number of kernels          & $\{3, 5, 7, \ldots, 15\}$        & 7       & 3       & 11      \\
\bottomrule
\end{tabular}
\label{table:DAN_hps}
\end{table}
\endgroup

\begingroup
\linespread{1.0}\selectfont
\begin{table}
\centering
\caption{Same as Table~\ref{table:ADDA_hps}, but using DANN.}
\begin{tabular}{l l l l l l}
\toprule
& & & \multicolumn{3}{c}{Best run} \\
\cmidrule(lr){4-6}
& & & sim-to-sim & sim-to-real & sim-to-real \\
Architecture & Parameter & Search Space & \HPGe & \LaBr & \NaI \\
\midrule
MLP & FE learning rate$^{\dagger}$ & $[10^{-8},\, 10^{-2}]_{\log}$ & 1.31e-6 & 1.05e-3 & 3.69e-4 \\
        & DD learning rate$^{\dagger}$ & $[10^{-8},\, 10^{-2}]_{\log}$ & 1.06e-5 & 1.52e-4 & 1.17e-4 \\
        & Batch size & $[64,\, 512]_{2^k}$ & 256 & 512 & 512 \\
        & Weight decay & $[10^{-7},\, 10^{-3}]_{\log}$ & 1.26e-7 & 2.06e-4 & 1.12e-6 \\
        & Dropout & $[0.0,\, 0.4]$ & 0.0588 & 0.046 & 8.29e-3 \\
        & Num GRL layers & $\{1, 2\}$ & 1 & 2 & 2 \\
        & GRL dense1 units & $[64,\, 2048]_{2^k}$ & 1024 & 2048 & 64 \\
        & GRL dense2 units & $[32,\, 1024]_{2^k}$ & N/A & 1024 & 32 \\
        & GRL $\kappa$ & $[0.0,\, 1.0]$ & 0.643 & 0.634 & 0.777 \\
\midrule
CNN & FE learning rate$^{\dagger}$ & $[10^{-8},\, 10^{-2}]_{\log}$ & 1.79e-3 & 1.24e-5 & 4.62e-5 \\
        & DD learning rate$^{\dagger}$ & $[10^{-8},\, 10^{-2}]_{\log}$ & 1.83e-4 & 4.29e-5 & 4.30e-4 \\
        & Batch size & $[64,\, 512]_{2^k}$ & 64 & 512 & 256 \\
        & Weight decay & $[10^{-7},\, 10^{-3}]_{\log}$ & 8.55e-7 & 4.61e-4 & 1.06e-5 \\
        & Dropout & $[0.0,\, 0.4]$ & 0.386 & 9.40e-3 & 0.223 \\
        & Num GRL layers & $\{1, 2\}$ & 2 & 2 & 1 \\
        & GRL dense1 units & $[64,\, 2048]_{2^k}$ & 2048 & 2048 & 64 \\
        & GRL dense2 units & $[32,\, 1024]_{2^k}$ & 1024 & 1024 & N/A \\
        & GRL $\kappa$ & $[0.0,\, 1.0]$ & 0.372 & 0.665 & 0.491 \\
\midrule
TBNN (Li \emph{et al.}) & FE learning rate$^{\dagger}$ & $[10^{-8},\, 10^{-2}]_{\log}$ & 4.24e-5 & 1.27e-3 & 2.21e-3 \\
        & DD learning rate$^{\dagger}$ & $[10^{-8},\, 10^{-2}]_{\log}$ & 7.52e-5 & 3.89e-4 & 4.52e-4 \\
        & Batch size & $[64,\, 512]_{2^k}$ & 64 & 128 & 512 \\
        & Weight decay & $[10^{-7},\, 10^{-3}]_{\log}$ & 2.93e-6 & 1.25e-5 & 2.41e-6 \\
        & Dropout & $[0.0,\, 0.4]$ & 0.0283 & 0.0215 & 0.328 \\
        & Num GRL layers & $\{1, 2\}$ & 1 & 2 & 2 \\
        & GRL dense1 units & $[64,\, 2048]_{2^k}$ & 128 & 512 & 256 \\
        & GRL dense2 units & $[32,\, 1024]_{2^k}$ & N/A & 256 & 128 \\
        & GRL $\kappa$ & $[0.0,\, 1.0]$ & 0.573 & 0.89 & 0.117 \\
\midrule
TBNN-LinEmb & FE learning rate$^{\dagger}$ & $[10^{-8},\, 10^{-2}]_{\log}$ & 4.81e-5 & 1.01e-4 & 2.31e-4 \\
        & DD learning rate$^{\dagger}$ & $[10^{-8},\, 10^{-2}]_{\log}$ & 4.41e-5 & 2.80e-4 & 1.62e-3 \\
        & Batch size & $[64,\, 512]_{2^k}$ & 512 & 256 & 256 \\
        & Weight decay & $[10^{-7},\, 10^{-3}]_{\log}$ & 1.99e-5 & 1.29e-4 & 5.97e-7 \\
        & Dropout & $[0.0,\, 0.4]$ & 0.139 & 0.0126 & 0.0986 \\
        & Num GRL layers & $\{1, 2\}$ & 2 & 2 & 2 \\
        & GRL dense1 units & $[64,\, 2048]_{2^k}$ & 1024 & 2048 & 512 \\
        & GRL dense2 units & $[32,\, 1024]_{2^k}$ & 512 & 1024 & 256 \\
        & GRL $\kappa$ & $[0.0,\, 1.0]$ & 0.0676 & 0.665 & 0.813 \\
\midrule
TBNN-NonLinEmb & FE learning rate$^{\dagger}$ & $[10^{-8},\, 10^{-2}]_{\log}$ & 5.41e-5 & 5.21e-5 & 3.26e-4 \\
        & DD learning rate$^{\dagger}$ & $[10^{-8},\, 10^{-2}]_{\log}$ & 2.76e-5 & 2.37e-4 & 8.01e-4 \\
        & Batch size & $[64,\, 512]_{2^k}$ & 512 & 64 & 512 \\
        & Weight decay & $[10^{-7},\, 10^{-3}]_{\log}$ & 1.54e-7 & 2.72e-5 & 1.18e-4 \\
        & Dropout & $[0.0,\, 0.4]$ & 0.193 & 0.162 & 0.123 \\
        & Num GRL layers & $\{1, 2\}$ & 2 & 2 & 2 \\
        & GRL dense1 units & $[64,\, 2048]_{2^k}$ & 1024 & 1024 & 2048 \\
        & GRL dense2 units & $[32,\, 1024]_{2^k}$ & 512 & 512 & 1024 \\
        & GRL $\kappa$ & $[0.0,\, 1.0]$ & 0.207 & 0.456 & 0.718 \\
\bottomrule
\end{tabular}
\par\smallskip
{\footnotesize $^{\dagger}$The FE and DD learning rates are jointly parameterized as $\ell_{\mathrm{FE}} = \ell/\sqrt{r}$ and $\ell_{\mathrm{DD}} = \ell\sqrt{r}$, where $\ell \in [10^{-7},\,10^{-3}]_{\log}$ is a shared base learning rate and $r \in [10^{-1},\,10^{1}]_{\log}$ is a ratio parameter. The intervals shown are the effective ranges.}
\label{table:DANN_hps}
\end{table}
\endgroup

\begingroup
\linespread{1.0}\selectfont
\begin{table}
\centering
\caption{Same as Table~\ref{table:ADDA_hps}, but using DeepCORAL.}
\begin{tabular}{l l l l l l}
\toprule
& & & \multicolumn{3}{c}{Best run} \\
\cmidrule(lr){4-6}
 & & & sim-to-sim & sim-to-real & sim-to-real \\
Architecture & Parameter & Search Space & \HPGe & \LaBr & \NaI \\
\midrule
MLP     & Learning rate              & $[10^{-7},\, 10^{-3}]_{\log}$   & 1.34e-4 & 8.95e-4 & 9.41e-4 \\
        & Batch size                 & $[64,\, 512]_{2^k}$      & 512     & 512     & 128     \\
        & Weight decay               & $[10^{-7},\, 10^{-3}]_{\log}$   & 1.37e-6 & 1.19e-6 & 6.11e-7 \\
        & Dropout                    & $[0.0,\, 0.4]$     & 0.297   & 8.24e-3 & 0.0839  \\
        & Tradeoff parameter         & $[10^{-1},\, 10^{10}]_{\log}$   & 7.13e4  & 1.20e9  & 5.77e7  \\
\midrule
CNN     & Learning rate              & $[10^{-7},\, 10^{-3}]_{\log}$   & 5.12e-4 & 2.97e-5 & 1.50e-5 \\
        & Batch size                 & $[64,\, 512]_{2^k}$      & 512     & 512     & 512     \\
        & Weight decay               & $[10^{-7},\, 10^{-3}]_{\log}$   & 2.21e-7 & 1.38e-7 & 1.90e-6 \\
        & Dropout                    & $[0.0,\, 0.4]$     & 0.036   & 5.71e-3 & 8.23e-3 \\
        & Tradeoff parameter         & $[10^{-1},\, 10^{10}]_{\log}$   & 2.82e5  & 1273    & 10.8    \\
\midrule
TBNN (Li \emph{et al.})
        & Learning rate              & $[10^{-7},\, 10^{-3}]_{\log}$   & 6.08e-4 & 7.82e-4 & 9.98e-4 \\
        & Batch size                 & $[64,\, 512]_{2^k}$      & 256     & 128     & 128     \\
        & Weight decay               & $[10^{-7},\, 10^{-3}]_{\log}$   & 5.70e-5 & 4.02e-4 & 1.83e-7 \\
        & Dropout                    & $[0.0,\, 0.4]$     & 5.46e-3 & 0.289   & 0.119   \\
        & Tradeoff parameter         & $[10^{-1},\, 10^{10}]_{\log}$   & 9.96e6  & 1.16e5  & 4.39e3  \\
\midrule
TBNN-LinEmb
        & Learning rate              & $[10^{-7},\, 10^{-3}]_{\log}$   & 3.69e-4 & 7.16e-4 & 4.75e-4 \\
        & Batch size                 & $[64,\, 512]_{2^k}$      & 512     & 512     & 512     \\
        & Weight decay               & $[10^{-7},\, 10^{-3}]_{\log}$   & 3.11e-4 & 1.21e-7 & 4.10e-6 \\
        & Dropout                    & $[0.0,\, 0.4]$     & 0.0566  & 0.228   & 0.0166  \\
        & Tradeoff parameter         & $[10^{-1},\, 10^{10}]_{\log}$   & 1.55e4  & 3.31e3  & 863     \\
\midrule
TBNN-NonLinEmb
        & Learning rate              & $[10^{-7},\, 10^{-3}]_{\log}$   & 1.24e-4 & 8.16e-4 & 5.21e-4 \\
        & Batch size                 & $[64,\, 512]_{2^k}$      & 512     & 512     & 512     \\
        & Weight decay               & $[10^{-7},\, 10^{-3}]_{\log}$   & 7.84e-7 & 9.77e-6 & 2.31e-5 \\
        & Dropout                    & $[0.0,\, 0.4]$     & 0.131   & 0.0632  & 0.221   \\
        & Tradeoff parameter         & $[10^{-1},\, 10^{10}]_{\log}$   & 1.85e4  & 3.78e3  & 3.65e3  \\
\bottomrule
\end{tabular}
\label{table:DeepCORAL_hps}
\end{table}
\endgroup

\begingroup
\linespread{1.0}\selectfont
\begin{table}
\centering
\caption{Same as Table~\ref{table:ADDA_hps}, but using DeepJDOT.}
\begin{tabular}{l l l l l l}
\toprule
& & & \multicolumn{3}{c}{Best run} \\
\cmidrule(lr){4-6}
 & & & sim-to-sim & sim-to-real & sim-to-real \\
Architecture & Parameter & Search Space & \HPGe & \LaBr & \NaI \\
\midrule
MLP     & Learning rate              & $[10^{-7},\, 10^{-3}]_{\log}$     & 9.25e-4 & 7.15e-4 & 7.24e-4 \\
        & Batch size                 & $[64,\, 512]_{2^k}$        & 512     & 512     & 256     \\
        & Weight decay               & $[10^{-7},\, 10^{-3}]_{\log}$     & 8.33e-6 & 5.23e-6 & 9.17e-4 \\
        & Dropout                    & $[0.0,\, 0.4]$       & 0.207   & 0.177   & 0.216   \\
        & Tradeoff parameter         & $[10^{-2},\, 10^{2}]_{\log}$      & 3.60    & 0.0204  & 0.0107  \\
        & Sinkhorn regularization    & $[10^{-2},\, 10^{1}]_{\log}$      & 8.47    & 4.36    & 7.57    \\
        & Sinkhorn iterations        & $\{5, \ldots, 30\}$          & 11      & 24      & 17      \\
        & JDOT $\alpha$              & $[10^{-2},\, 10^{1}]_{\log}$      & 0.397   & 0.205   & 1.14    \\
        & JDOT $\beta$               & $[0.0,\, 2.0]$       & 1.36    & 0.222   & 1.41    \\
\midrule
CNN     & Learning rate              & $[10^{-7},\, 10^{-3}]_{\log}$     & 7.18e-4 & 1.33e-6 & 9.03e-4 \\
        & Batch size                 & $[64,\, 512]_{2^k}$        & 64      & 512     & 128     \\
        & Weight decay               & $[10^{-7},\, 10^{-3}]_{\log}$     & 8.69e-4 & 1.02e-4 & 7.92e-7 \\
        & Dropout                    & $[0.0,\, 0.4]$       & 0.383   & 0.100   & 0.139   \\
        & Tradeoff parameter         & $[10^{-2},\, 10^{2}]_{\log}$      & 0.0754  & 2.24    & 1.31    \\
        & Sinkhorn regularization    & $[10^{-2},\, 10^{1}]_{\log}$      & 0.0364  & 1.85    & 0.0446  \\
        & Sinkhorn iterations        & $\{5, \ldots, 30\}$          & 5       & 17      & 19      \\
        & JDOT $\alpha$              & $[10^{-2},\, 10^{1}]_{\log}$      & 1.005   & 0.0167  & 0.0127  \\
        & JDOT $\beta$               & $[0.0,\, 2.0]$       & 0.499   & 1.31    & 1.16    \\
\midrule
TBNN (Li \emph{et al.})
        & Learning rate              & $[10^{-7},\, 10^{-3}]_{\log}$     & 8.97e-5 & 2.25e-4 & 8.75e-4 \\
        & Batch size                 & $[64,\, 512]_{2^k}$        & 512     & 128     & 128     \\
        & Weight decay               & $[10^{-7},\, 10^{-3}]_{\log}$     & 3.57e-7 & 4.17e-5 & 1.56e-5 \\
        & Dropout                    & $[0.0,\, 0.4]$       & 0.0817  & 0.106   & 0.111   \\
        & Tradeoff parameter         & $[10^{-2},\, 10^{2}]_{\log}$      & 19.1    & 5.06    & 0.730   \\
        & Sinkhorn regularization    & $[10^{-2},\, 10^{1}]_{\log}$      & 0.0436  & 0.242   & 0.0132  \\
        & Sinkhorn iterations        & $\{5, \ldots, 30\}$          & 7       & 19      & 5       \\
        & JDOT $\alpha$              & $[10^{-2},\, 10^{1}]_{\log}$      & 0.0569  & 0.0126  & 0.0549  \\
        & JDOT $\beta$               & $[0.0,\, 2.0]$       & 1.21    & 1.12    & 0.413   \\
\midrule
TBNN-LinEmb
        & Learning rate              & $[10^{-7},\, 10^{-3}]_{\log}$     & 2.81e-5 & 3.33e-5 & 1.94e-4 \\
        & Batch size                 & $[64,\, 512]_{2^k}$        & 64      & 64      & 128     \\
        & Weight decay               & $[10^{-7},\, 10^{-3}]_{\log}$     & 1.24e-6 & 7.67e-5 & 4.45e-4 \\
        & Dropout                    & $[0.0,\, 0.4]$       & 0.190   & 0.231   & 0.376   \\
        & Tradeoff parameter         & $[10^{-2},\, 10^{2}]_{\log}$      & 0.343   & 68.9    & 0.0533  \\
        & Sinkhorn regularization    & $[10^{-2},\, 10^{1}]_{\log}$      & 0.753   & 0.707   & 2.11    \\
        & Sinkhorn iterations        & $\{5, \ldots, 30\}$          & 19      & 28      & 6       \\
        & JDOT $\alpha$              & $[10^{-2},\, 10^{1}]_{\log}$      & 4.57    & 0.0119  & 0.239   \\
        & JDOT $\beta$               & $[0.0,\, 2.0]$       & 0.532   & 1.01    & 1.89    \\
\midrule
TBNN-NonLinEmb
        & Learning rate              & $[10^{-7},\, 10^{-3}]_{\log}$     & 7.54e-4 & 3.17e-6 & 1.52e-4 \\
        & Batch size                 & $[64,\, 512]_{2^k}$        & 64      & 64      & 256     \\
        & Weight decay               & $[10^{-7},\, 10^{-3}]_{\log}$     & 5.23e-7 & 6.04e-5 & 1.80e-5 \\
        & Dropout                    & $[0.0,\, 0.4]$       & 0.239   & 0.260   & 0.0582  \\
        & Tradeoff parameter         & $[10^{-2},\, 10^{2}]_{\log}$      & 1.17    & 0.0562  & 0.0134  \\
        & Sinkhorn regularization    & $[10^{-2},\, 10^{1}]_{\log}$      & 0.0401  & 1.38    & 1.63    \\
        & Sinkhorn iterations        & $\{5, \ldots, 30\}$          & 16      & 24      & 6       \\
        & JDOT $\alpha$              & $[10^{-2},\, 10^{1}]_{\log}$      & 0.141   & 0.0424  & 0.0902  \\
        & JDOT $\beta$               & $[0.0,\, 2.0]$       & 0.303   & 1.42    & 0.794   \\
\bottomrule
\end{tabular}
\label{table:DeepJDOT_hps}
\end{table}
\endgroup

\begingroup
\linespread{1.0}\selectfont
\begin{table}
\centering
\caption{Same as Table~\ref{table:ADDA_hps}, but using Mean Teacher.}
\begin{tabular}{l l l l l l}
\toprule
& & & \multicolumn{3}{c}{Best run} \\
\cmidrule(lr){4-6}
 & & & sim-to-sim & sim-to-real & sim-to-real \\
Architecture & Parameter & Search Space & \HPGe & \LaBr & \NaI \\
\midrule
MLP     & Learning rate         & $[10^{-7},\, 10^{-3}]_{\log}$     & 9.06e-4 & 5.36e-4 & 8.24e-4 \\
        & Batch size            & $[64,\, 512]_{2^k}$        & 512     & 64      & 64      \\
        & Weight decay          & $[10^{-7},\, 10^{-3}]_{\log}$     & 7.45e-7 & 6.91e-6 & 6.83e-6 \\
        & Dropout               & $[0.0,\, 0.4]$       & 0.0229  & 0.183   & 0.121   \\
        & Tradeoff parameter    & $[10^{-2},\, 10^{3}]_{\log}$      & 689     & 430     & 273     \\
        & EMA decay             & $[0.95,\, 0.9999]$    & 0.9969  & 0.9559  & 0.9520  \\
        & Effective counts      & $[10^{2},\, 5{\times}10^{4}]_{\log}$       & 3.91e4  & 213     & 169     \\
\midrule
CNN     & Learning rate         & $[10^{-7},\, 10^{-3}]_{\log}$     & 7.57e-4 & 2.42e-4 & 7.16e-4 \\
        & Batch size            & $[64,\, 512]_{2^k}$        & 64      & 256     & 128     \\
        & Weight decay          & $[10^{-7},\, 10^{-3}]_{\log}$     & 7.36e-4 & 7.51e-7 & 2.51e-5 \\
        & Dropout               & $[0.0,\, 0.4]$       & 0.365   & 0.0831  & 0.146   \\
        & Tradeoff parameter    & $[10^{-2},\, 10^{3}]_{\log}$      & 10.1    & 818     & 722     \\
        & EMA decay             & $[0.95,\, 0.9999]$    & 0.9928  & 0.9634  & 0.9504  \\
        & Effective counts      & $[10^{2},\, 5{\times}10^{4}]_{\log}$       & 4.22e3  & 257     & 124     \\
\midrule
TBNN (Li \emph{et al.})
        & Learning rate         & $[10^{-7},\, 10^{-3}]_{\log}$     & 3.36e-4 & 9.62e-4 & 7.52e-4 \\
        & Batch size            & $[64,\, 512]_{2^k}$        & 256     & 64      & 128     \\
        & Weight decay          & $[10^{-7},\, 10^{-3}]_{\log}$     & 8.01e-5 & 2.77e-4 & 1.84e-4 \\
        & Dropout               & $[0.0,\, 0.4]$       & 0.0461  & 0.363   & 0.283   \\
        & Tradeoff parameter    & $[10^{-2},\, 10^{3}]_{\log}$      & 63.4    & 3.02    & 0.558   \\
        & EMA decay             & $[0.95,\, 0.9999]$    & 0.9723  & 0.9523  & 0.9783  \\
        & Effective counts      & $[10^{2},\, 5{\times}10^{4}]_{\log}$       & 1.44e4  & 3.12e4  & 6.29e3  \\
\midrule
TBNN-LinEmb
        & Learning rate         & $[10^{-7},\, 10^{-3}]_{\log}$     & 8.51e-4 & 9.26e-4 & 8.64e-4 \\
        & Batch size            & $[64,\, 512]_{2^k}$        & 512     & 128     & 64      \\
        & Weight decay          & $[10^{-7},\, 10^{-3}]_{\log}$     & 7.26e-6 & 7.13e-7 & 1.22e-4 \\
        & Dropout               & $[0.0,\, 0.4]$       & 0.0878  & 0.0149  & 9.67e-3 \\
        & Tradeoff parameter    & $[10^{-2},\, 10^{3}]_{\log}$      & 8.58    & 322     & 3.83    \\
        & EMA decay             & $[0.95,\, 0.9999]$    & 0.9814  & 0.9942  & 0.9526  \\
        & Effective counts      & $[10^{2},\, 5{\times}10^{4}]_{\log}$       & 3.92e4  & 395     & 224     \\
\midrule
TBNN-NonLinEmb
        & Learning rate         & $[10^{-7},\, 10^{-3}]_{\log}$     & 7.93e-4 & 8.15e-4 & 5.58e-4 \\
        & Batch size            & $[64,\, 512]_{2^k}$        & 128     & 64      & 64      \\
        & Weight decay          & $[10^{-7},\, 10^{-3}]_{\log}$     & 1.76e-7 & 2.03e-6 & 2.36e-6 \\
        & Dropout               & $[0.0,\, 0.4]$       & 0.0317  & 0.0454  & 0.0108  \\
        & Tradeoff parameter    & $[10^{-2},\, 10^{3}]_{\log}$      & 3.76    & 2.17    & 1.92    \\
        & EMA decay             & $[0.95,\, 0.9999]$    & 0.9956  & 0.9591  & 0.9571  \\
        & Effective counts      & $[10^{2},\, 5{\times}10^{4}]_{\log}$       & 5.39e3  & 172     & 336     \\
\bottomrule
\end{tabular}
\label{table:MeanTeacher_hps}
\end{table}
\endgroup

\begingroup
\linespread{1.0}\selectfont
\begin{table}
\centering
\caption{Same as Table~\ref{table:ADDA_hps}, but using SimCLR.}
\begin{tabular}{l l l l l l}
\toprule
& & & \multicolumn{3}{c}{Best run} \\
\cmidrule(lr){4-6}
 & & & sim-to-sim & sim-to-real & sim-to-real \\
Architecture & Parameter & Search Space & \HPGe & \LaBr & \NaI \\
\midrule
MLP     & Learning rate          & $[10^{-7},\, 10^{-3}]_{\log}$     & 5.85e-4 & 3.21e-4 & 6.21e-4 \\
        & Batch size             & $[64,\, 512]_{2^k}$        & 128     & 128     & 128     \\
        & Weight decay           & $[10^{-7},\, 10^{-3}]_{\log}$     & 2.00e-5 & 1.99e-4 & 5.92e-7 \\
        & Dropout                & $[0.0,\, 0.4]$       & 0.124   & 0.101   & 0.311   \\
        & Tradeoff parameter     & $[10^{-2},\, 10^{3}]_{\log}$      & 670     & 41.2    & 74.8    \\
        & Temperature            & $[10^{-2},\, 0.5]_{\log}$      & 0.0918  & 0.499   & 0.491   \\
        & Projection layers      & $\{1, 2, 3\}$           & 1       & 1       & 1       \\
        & Projection width       & $[64,\, 1024]_{2^k}$       & 128     & 256     & 256     \\
        & Effective counts       & $[10^{2},\, 10^{5}]_{\log}$       & 3.02e4  & 4.69e3  & 1.84e3  \\
\midrule
CNN     & Learning rate          & $[10^{-7},\, 10^{-3}]_{\log}$     & 1.40e-4 & 5.82e-6 & 9.08e-4 \\
        & Batch size             & $[64,\, 512]_{2^k}$        & 512     & 256     & 256     \\
        & Weight decay           & $[10^{-7},\, 10^{-3}]_{\log}$     & 4.98e-7 & 6.44e-5 & 4.37e-6 \\
        & Dropout                & $[0.0,\, 0.4]$       & 0.0990  & 0.262   & 0.103   \\
        & Tradeoff parameter     & $[10^{-2},\, 10^{3}]_{\log}$      & 95.7    & 472     & 2.63    \\
        & Temperature            & $[10^{-2},\, 0.5]_{\log}$      & 0.0664  & 0.173   & 0.449   \\
        & Projection layers      & $\{1, 2, 3\}$           & 1       & 2       & 1       \\
        & Projection width       & $[64,\, 1024]_{2^k}$       & 1024    & 1024    & 1024    \\
        & Effective counts       & $[10^{2},\, 10^{5}]_{\log}$       & 7.42e4  & 1.38e4  & 8.40e4  \\
\midrule
TBNN (Li \emph{et al.})
        & Learning rate          & $[10^{-7},\, 10^{-3}]_{\log}$     & 6.80e-4 & 8.78e-4 & 7.01e-4 \\
        & Batch size             & $[64,\, 512]_{2^k}$        & 512     & 64      & 128     \\
        & Weight decay           & $[10^{-7},\, 10^{-3}]_{\log}$     & 5.44e-7 & 4.21e-7 & 5.65e-7 \\
        & Dropout                & $[0.0,\, 0.4]$       & 0.0500  & 0.283   & 0.266   \\
        & Tradeoff parameter     & $[10^{-2},\, 10^{3}]_{\log}$      & 0.151   & 362     & 1.61    \\
        & Temperature            & $[10^{-2},\, 0.5]_{\log}$      & 0.0364  & 0.154   & 0.0821  \\
        & Projection layers      & $\{1, 2, 3\}$           & 2       & 1       & 1       \\
        & Projection width       & $[64,\, 1024]_{2^k}$       & 128     & 128     & 128     \\
        & Effective counts       & $[10^{2},\, 10^{5}]_{\log}$       & 2.29e4  & 5.98e3  & 7.16e4  \\
\midrule
TBNN-LinEmb
        & Learning rate          & $[10^{-7},\, 10^{-3}]_{\log}$     & 9.58e-5 & 9.24e-4 & 4.58e-4 \\
        & Batch size             & $[64,\, 512]_{2^k}$        & 128     & 128     & 64      \\
        & Weight decay           & $[10^{-7},\, 10^{-3}]_{\log}$     & 3.62e-7 & 1.61e-6 & 7.28e-6 \\
        & Dropout                & $[0.0,\, 0.4]$       & 0.208   & 0.159   & 0.142   \\
        & Tradeoff parameter     & $[10^{-2},\, 10^{3}]_{\log}$      & 0.418   & 1.55    & 278     \\
        & Temperature            & $[10^{-2},\, 0.5]_{\log}$      & 0.0469  & 0.201   & 0.443   \\
        & Projection layers      & $\{1, 2, 3\}$           & 1       & 1       & 2       \\
        & Projection width       & $[64,\, 1024]_{2^k}$       & 1024    & 64      & 128     \\
        & Effective counts       & $[10^{2},\, 10^{5}]_{\log}$       & 169     & 3.97e3  & 3.93e4  \\
\midrule
TBNN-NonLinEmb
        & Learning rate          & $[10^{-7},\, 10^{-3}]_{\log}$     & 9.06e-4 & 5.22e-5 & 7.50e-4 \\
        & Batch size             & $[64,\, 512]_{2^k}$        & 512     & 64      & 128     \\
        & Weight decay           & $[10^{-7},\, 10^{-3}]_{\log}$     & 3.58e-7 & 3.57e-6 & 9.72e-7 \\
        & Dropout                & $[0.0,\, 0.4]$       & 0.0785  & 0.166   & 0.207   \\
        & Tradeoff parameter     & $[10^{-2},\, 10^{3}]_{\log}$      & 36.8    & 13.2    & 177     \\
        & Temperature            & $[10^{-2},\, 0.5]_{\log}$      & 0.0738  & 0.430   & 0.495   \\
        & Projection layers      & $\{1, 2, 3\}$           & 1       & 1       & 3       \\
        & Projection width       & $[64,\, 1024]_{2^k}$       & 64      & 256     & 1024    \\
        & Effective counts       & $[10^{2},\, 10^{5}]_{\log}$       & 6.87e3  & 1.34e4  & 1.29e4  \\
\bottomrule
\end{tabular}
\label{table:SimCLR_hps}
\end{table}
\endgroup

\begin{table}
\centering
\caption{Calculated $p$-values from a one-sided Wilcoxon signed-rank test, testing the null hypothesis that UDA provides no improvement over the source-only baseline model. $p$-values below a significance threshold of $0.01$ are bolded, indicating statistically significant improvement.}
\label{table:UDA_pvalues}
\begin{subtable}{\textwidth}
\centering
\caption{Scenario: sim-to-sim (\HPGe). Statistical significance tests compare testing APE score (UDA $>$ source-only).}
\label{table:UDA_pvalues_hpge}
\begin{tabular}{l c c c c c}
\toprule
 & MLP & CNN & TBNN (Li \emph{et al.}) & TBNN-LinEmb & TBNN-NonlinEmb \\
\midrule
ADDA           & 1.000  & 0.995  & \textbf{0.001} & \textbf{0.001} & \textbf{0.002} \\
DAN            & 1.000  & 1.000  & \textbf{0.001} & \textbf{0.001} & \textbf{0.001} \\
DANN           & \textbf{0.001}  & 0.903  & \textbf{0.001}  & \textbf{0.001}  & \textbf{0.001} \\
DeepCORAL      & \textbf{0.001} & 1.000  & \textbf{0.007} & 0.998  & 1.000 \\
DeepJDOT       & 1.000  & 1.000  & 0.246  & 1.000  & 1.000 \\
Mean Teacher    & 1.000  & 1.000  & 0.216  & 1.000  & 1.000 \\
SimCLR         & 1.000  & 1.000  & \textbf{0.001} & 1.000  & 0.999 \\
\bottomrule
\end{tabular}
\end{subtable}

\vspace{1.5em}
\begin{subtable}{\textwidth}
\centering
\caption{Scenario: sim-to-real (\LaBr). Statistical significance tests compare testing accuracy (UDA $>$ source-only).}
\label{table:UDA_pvalues_labr}
\begin{tabular}{l c c c c c}
\toprule
 & MLP & CNN & TBNN (Li \emph{et al.}) & TBNN-LinEmb & TBNN-NonlinEmb \\
\midrule
ADDA           & \textbf{0.001} & \textbf{0.001} & \textbf{0.001} & \textbf{0.002} & \textbf{0.001} \\
DAN            & \textbf{0.001} & \textbf{0.002} & \textbf{0.001} & \textbf{0.001} & \textbf{0.001} \\
DANN           & \textbf{0.001}  & \textbf{0.001}  & \textbf{0.001}  & \textbf{0.001} & \textbf{0.001} \\
DeepCORAL      & \textbf{0.001} & \textbf{0.001} & \textbf{0.001} & \textbf{0.001} & \textbf{0.001} \\
DeepJDOT       & 0.042  & 0.010  & \textbf{0.005} & \textbf{0.001} & \textbf{0.001} \\
Mean Teacher   & \textbf{0.001} & 0.976  & \textbf{0.001} & 0.968  & 0.014 \\
SimCLR         & \textbf{0.001} & \textbf{0.001} & \textbf{0.001} & \textbf{0.001} & \textbf{0.001} \\
\bottomrule
\end{tabular}
\end{subtable}

\vspace{1.5em}
\begin{subtable}{\textwidth}
\centering
\caption{Scenario: sim-to-real (\NaI). Statistical significance tests compare testing accuracy (UDA $>$ source-only).}
\label{table:UDA_pvalues_nai}
\begin{tabular}{l c c c c c}
\toprule
 & MLP & CNN & TBNN (Li \emph{et al.}) & TBNN-LinEmb & TBNN-NonlinEmb \\
\midrule
ADDA           & \textbf{0.001} & 0.065  & \textbf{0.001} & \textbf{0.001} & 0.500 \\
DAN            & \textbf{0.001} & 0.188  & \textbf{0.001} & \textbf{0.001} & \textbf{0.001} \\
DANN           & \textbf{0.001} & 0.024  & \textbf{0.001} & \textbf{0.001} & \textbf{0.001} \\
DeepCORAL      & \textbf{0.001} & \textbf{0.001} & \textbf{0.001} & \textbf{0.001} & \textbf{0.001} \\
DeepJDOT       & \textbf{0.001} & 0.884  & \textbf{0.001} & \textbf{0.001} & \textbf{0.001} \\
Mean Teacher   & 0.991  & 0.999  & \textbf{0.001} & 0.019  & \textbf{0.001} \\
SimCLR         & \textbf{0.001} & \textbf{0.003} & \textbf{0.001} & \textbf{0.001} & \textbf{0.001} \\
\bottomrule
\end{tabular}
\end{subtable}
\end{table}

\begin{figure}
    \centering

    \begin{subfigure}[t]{0.38\textwidth}
        \centering
        \includegraphics[width=\textwidth]{UDA_umap_TBNN_linear_gap_source.pdf}
        \caption{Sim-to-sim \HPGe{} scenario, source-only model.}
        \label{fig:umap_hpge_source_appendix}
    \end{subfigure}
    \hfill
    \begin{subfigure}[t]{0.38\textwidth}
        \centering
        \includegraphics[width=\textwidth]{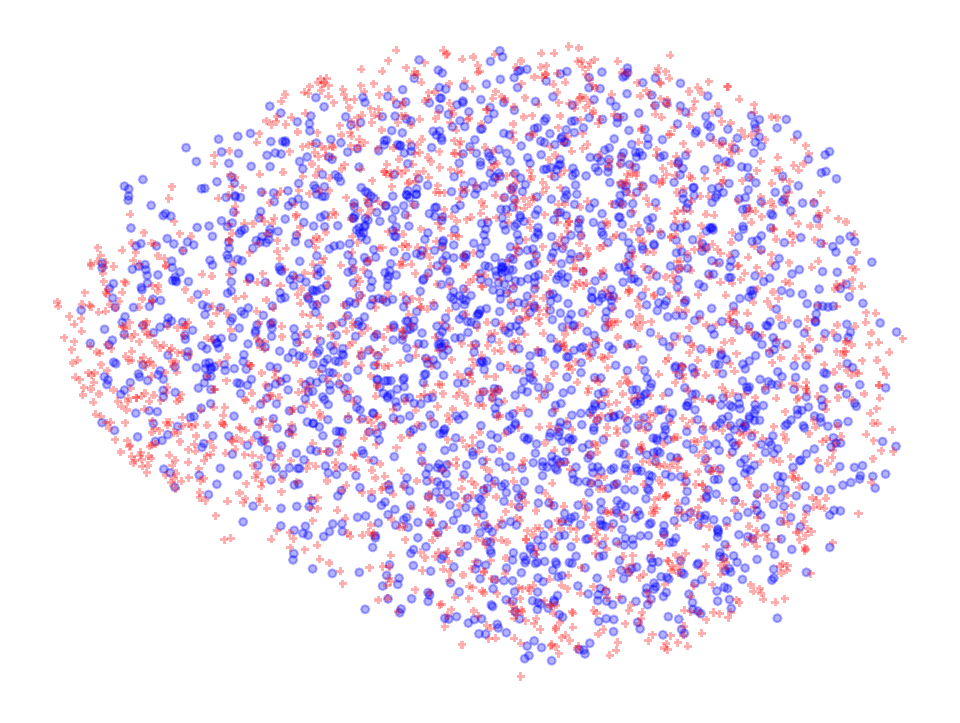}
        \caption{Sim-to-sim \HPGe{} scenario, ADDA model.}
        \label{fig:umap_hpge_ADDA}
    \end{subfigure}

    \vspace{0.5em}
    \begin{subfigure}[t]{0.38\textwidth}
        \centering
        \includegraphics[width=\textwidth]{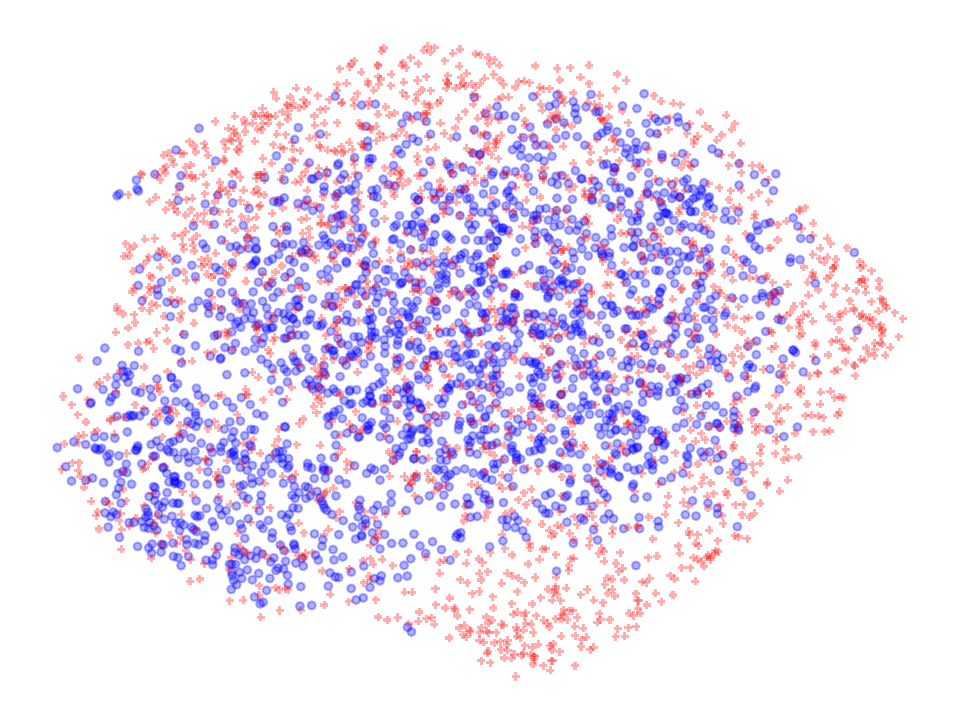}
        \caption{Sim-to-sim \HPGe{} scenario, DAN model.}
        \label{fig:umap_hpge_DAN}
    \end{subfigure}
    \hfill
    \begin{subfigure}[t]{0.38\textwidth}
        \centering
        \includegraphics[width=\textwidth]{UDA_umap_TBNN_linear_gap_DANN.pdf}
        \caption{Sim-to-sim \HPGe{} scenario, DANN model.}
        \label{fig:umap_hpge_DANN}
    \end{subfigure}

    \vspace{0.5em}
    \begin{subfigure}[t]{0.38\textwidth}
        \centering
        \includegraphics[width=\textwidth]{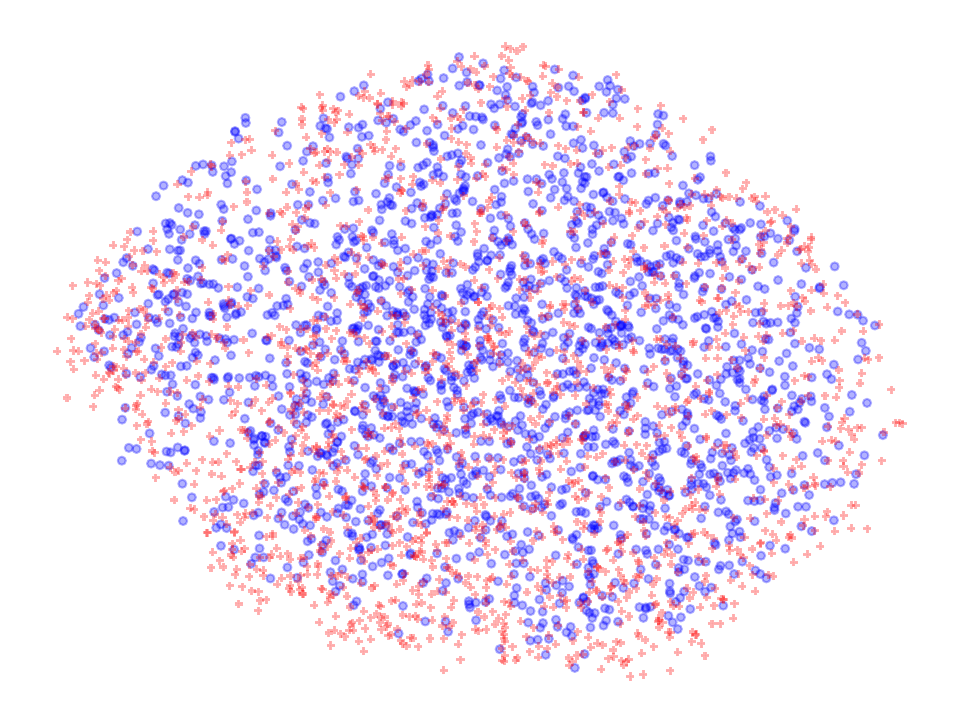}
        \caption{Sim-to-sim \HPGe{} scenario, DeepCORAL model.}
        \label{fig:umap_hpge_DeepCORAL}
    \end{subfigure}
    \hfill
    \begin{subfigure}[t]{0.38\textwidth}
        \centering
        \includegraphics[width=\textwidth]{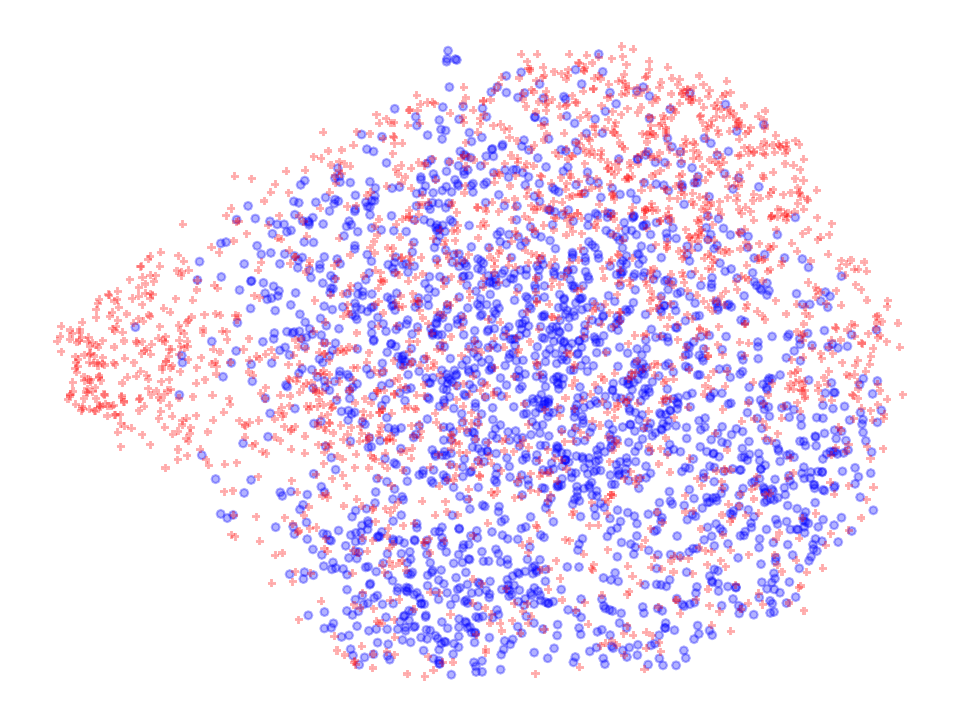}
        \caption{Sim-to-sim \HPGe{} scenario, DeepJDOT model.}
        \label{fig:umap_hpge_DeepJDOT}
    \end{subfigure}

    \vspace{0.5em}
    \begin{subfigure}[t]{0.38\textwidth}
        \centering
        \includegraphics[width=\textwidth]{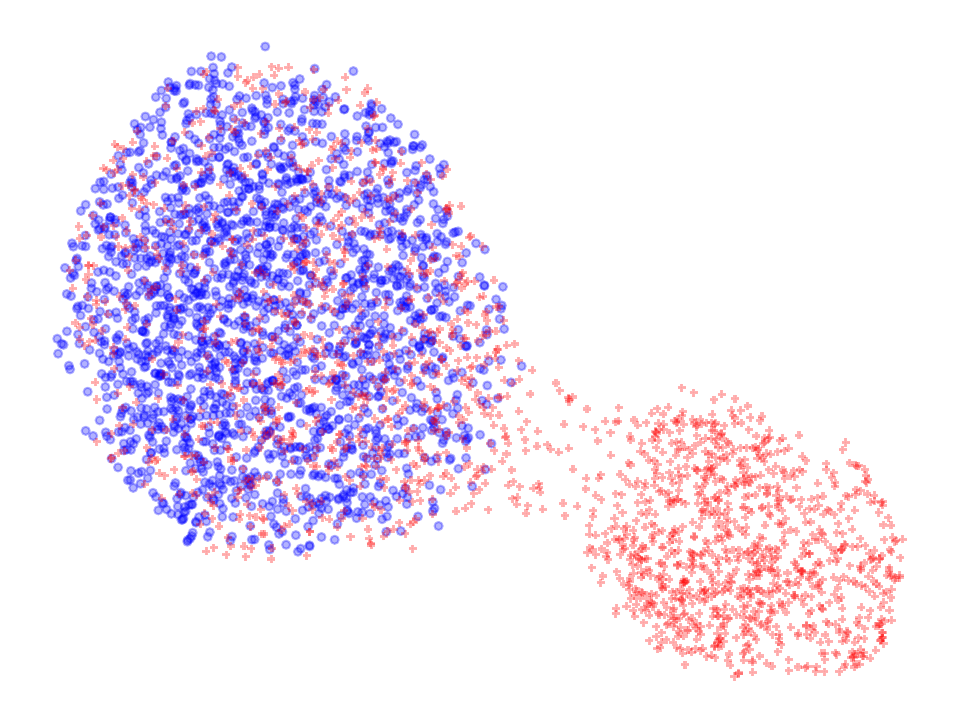}
        \caption{Sim-to-sim \HPGe{} scenario, Mean Teacher model.}
        \label{fig:umap_hpge_MeanTeacher}
    \end{subfigure}
    \hfill
    \begin{subfigure}[t]{0.38\textwidth}
        \centering
        \includegraphics[width=\textwidth]{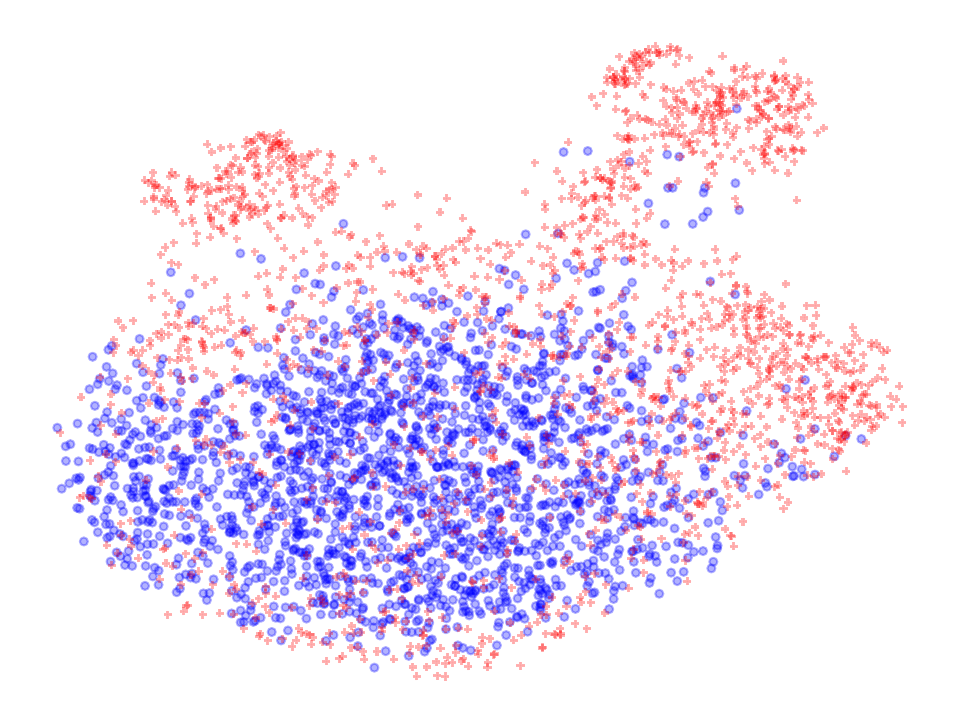}
        \caption{Sim-to-sim \HPGe{} scenario, SimCLR model.}
        \label{fig:umap_hpge_SimCLR}
    \end{subfigure}

    \caption{UMAP visualizations of the feature extractor outputs for source spectra (`\texttt{o}' markers) and target spectra (`\texttt{+}' markers) for all UDA methods for the sim-to-sim domain adaptation scenario. Color indicates domain (blue = source, red = target).}
    \label{fig:umap_hpge_all}
\end{figure}

\begin{figure}
    \centering

    \begin{subfigure}[t]{0.38\textwidth}
        \centering
        \includegraphics[width=\textwidth]{UDA_umap_TBNN_linear_gap_source_LaBr.pdf}
        \caption{Sim-to-real \LaBr{} scenario, source-only model.}
        \label{fig:umap_labr_source_appendix}
    \end{subfigure}
    \hfill
    \begin{subfigure}[t]{0.38\textwidth}
        \centering
        \includegraphics[width=\textwidth]{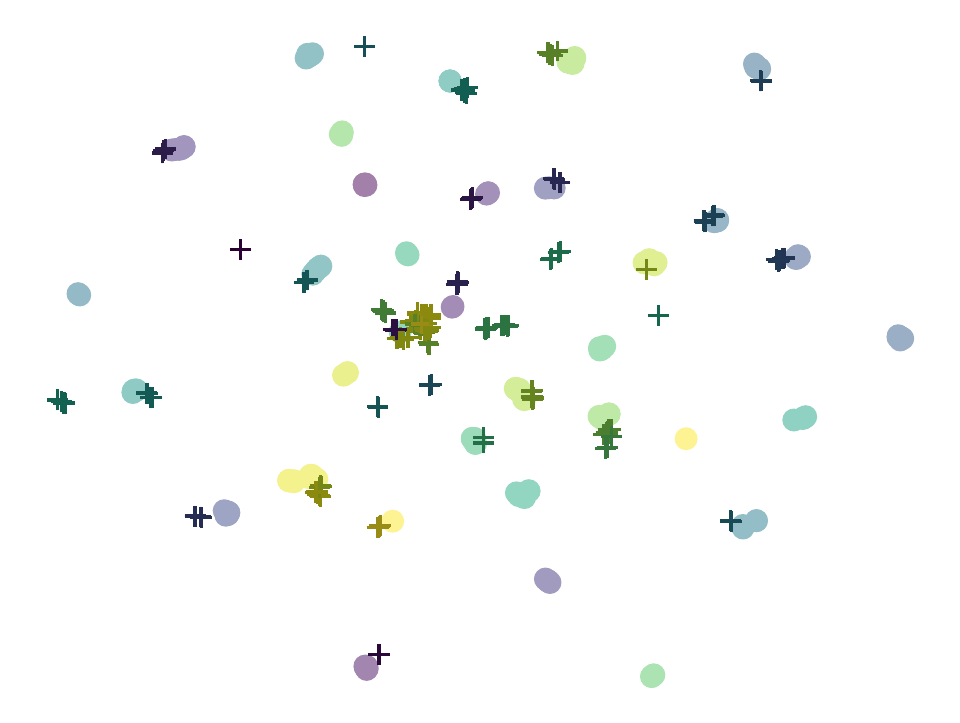}
        \caption{Sim-to-real \LaBr{} scenario, ADDA model.}
        \label{fig:umap_labr_ADDA}
    \end{subfigure}

    \vspace{0.5em}
    \begin{subfigure}[t]{0.38\textwidth}
        \centering
        \includegraphics[width=\textwidth]{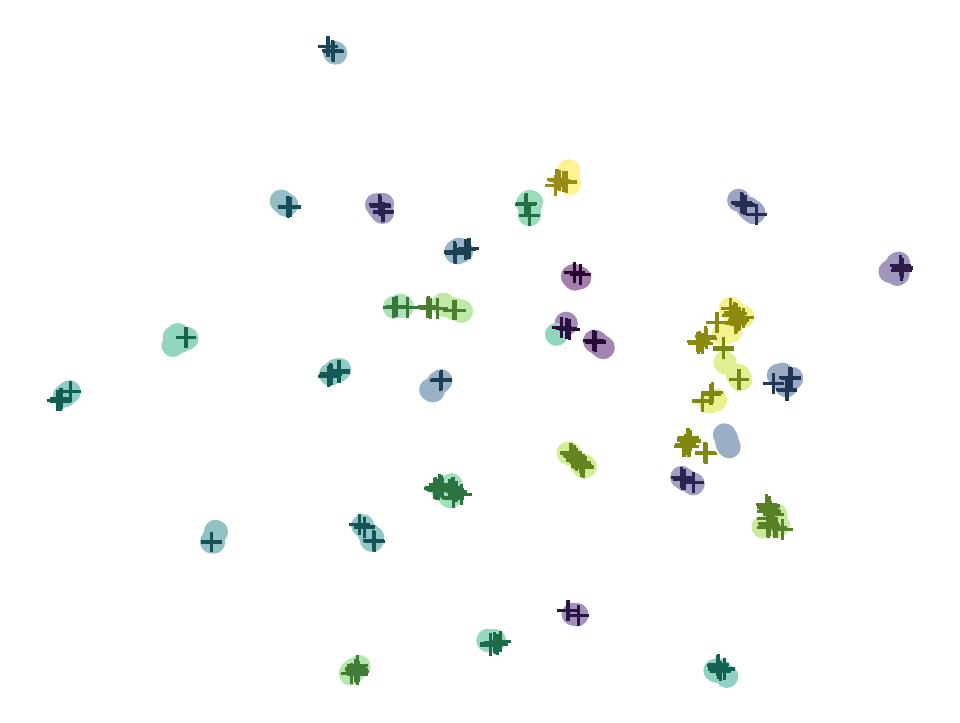}
        \caption{Sim-to-real \LaBr{} scenario, DAN model.}
        \label{fig:umap_labr_DAN}
    \end{subfigure}
    \hfill
    \begin{subfigure}[t]{0.38\textwidth}
        \centering
        \includegraphics[width=\textwidth]{UDA_umap_TBNN_linear_gap_DANN_LaBr.pdf}
        \caption{Sim-to-real \LaBr{} scenario, DANN model.}
        \label{fig:umap_labr_DANN}
    \end{subfigure}

    \vspace{0.5em}
    \begin{subfigure}[t]{0.38\textwidth}
        \centering
        \includegraphics[width=\textwidth]{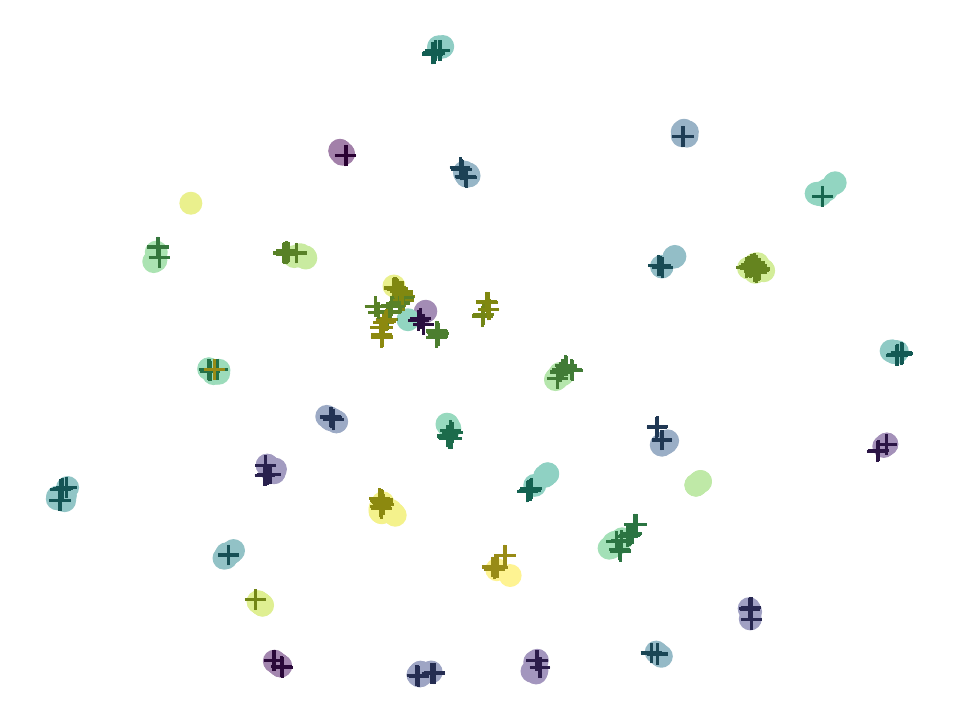}
        \caption{Sim-to-real \LaBr{} scenario, DeepCORAL model.}
        \label{fig:umap_labr_DeepCORAL}
    \end{subfigure}
    \hfill
    \begin{subfigure}[t]{0.38\textwidth}
        \centering
        \includegraphics[width=\textwidth]{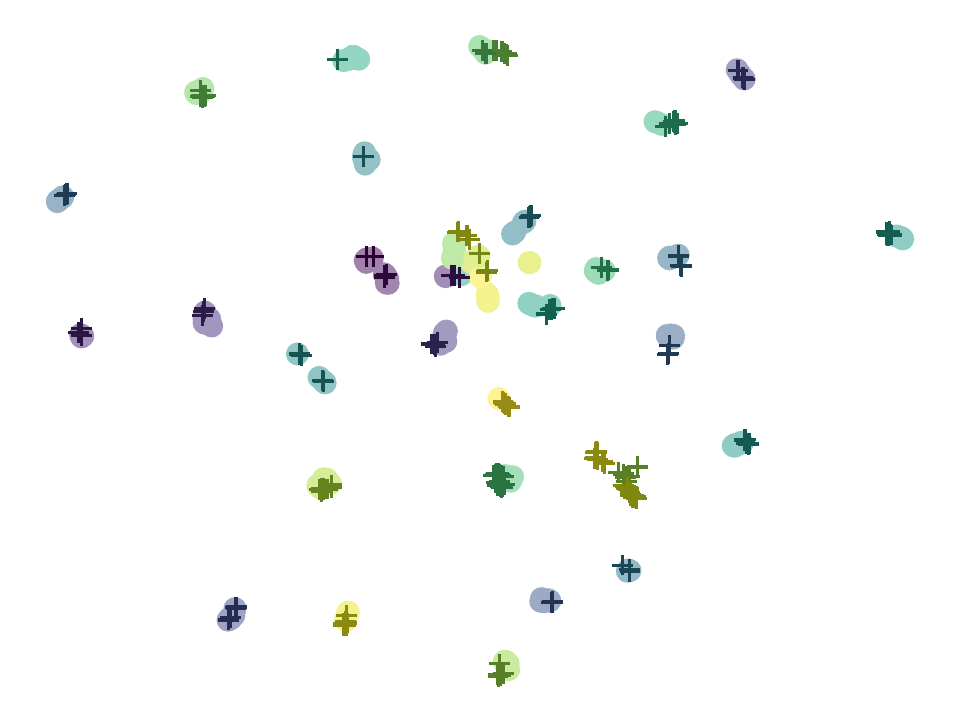}
        \caption{Sim-to-real \LaBr{} scenario, DeepJDOT model.}
        \label{fig:umap_labr_DeepJDOT}
    \end{subfigure}

    \vspace{0.5em}
    \begin{subfigure}[t]{0.38\textwidth}
        \centering
        \includegraphics[width=\textwidth]{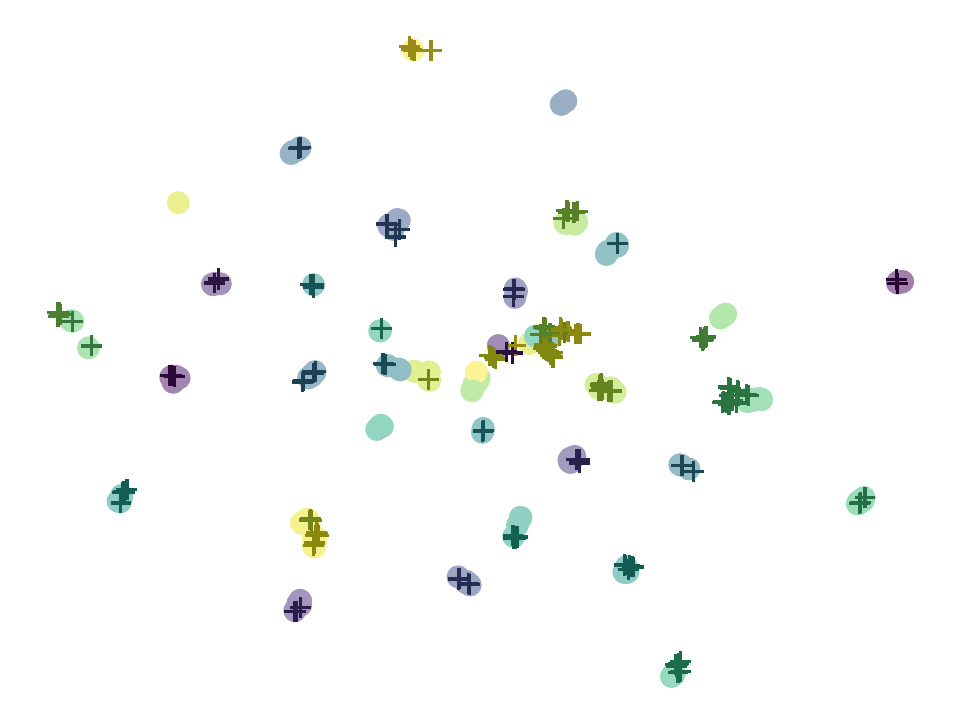}
        \caption{Sim-to-real \LaBr{} scenario, Mean Teacher model.}
        \label{fig:umap_labr_MeanTeacher}
    \end{subfigure}
    \hfill
    \begin{subfigure}[t]{0.38\textwidth}
        \centering
        \includegraphics[width=\textwidth]{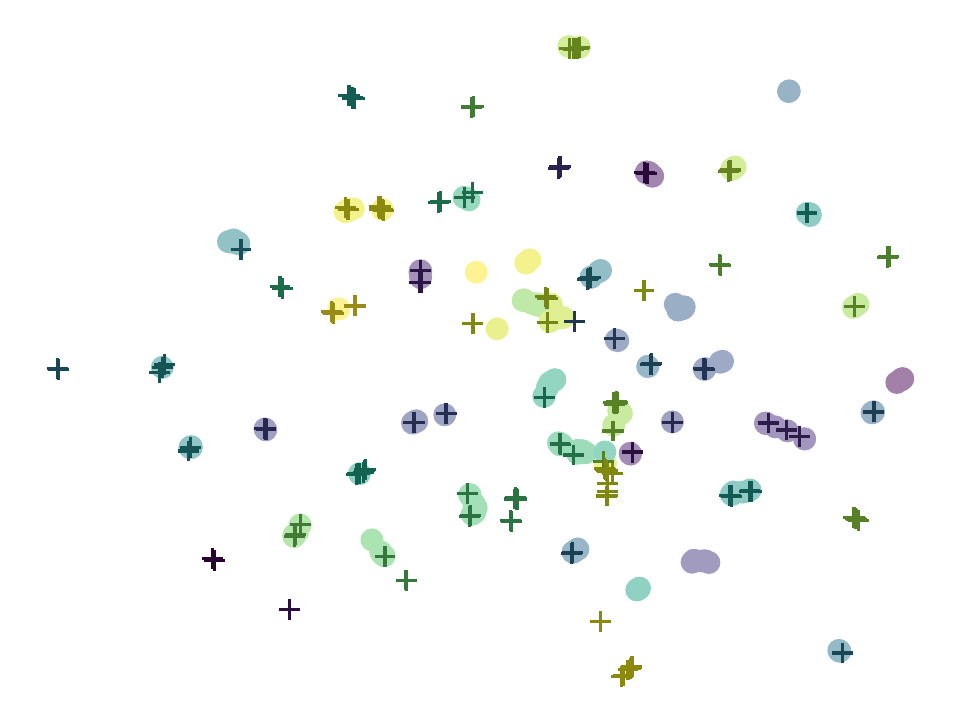}
        \caption{Sim-to-real \LaBr{} scenario, SimCLR model.}
        \label{fig:umap_labr_SimCLR}
    \end{subfigure}

    \caption{Same as Fig.~\ref{fig:umap_hpge_all}, but for the sim-to-real \LaBr{} domain adaptation scenario. Color indicates class.}
    \label{fig:umap_labr_all}
\end{figure}

\begin{figure}
    \centering

    \begin{subfigure}[t]{0.38\textwidth}
        \centering
        \includegraphics[width=\textwidth]{UDA_umap_TBNN_linear_gap_source_NaI.pdf}
        \caption{Sim-to-real \NaI{} scenario, source-only model.}
        \label{fig:umap_nai_source_appendix}
    \end{subfigure}
    \hfill
    \begin{subfigure}[t]{0.38\textwidth}
        \centering
        \includegraphics[width=\textwidth]{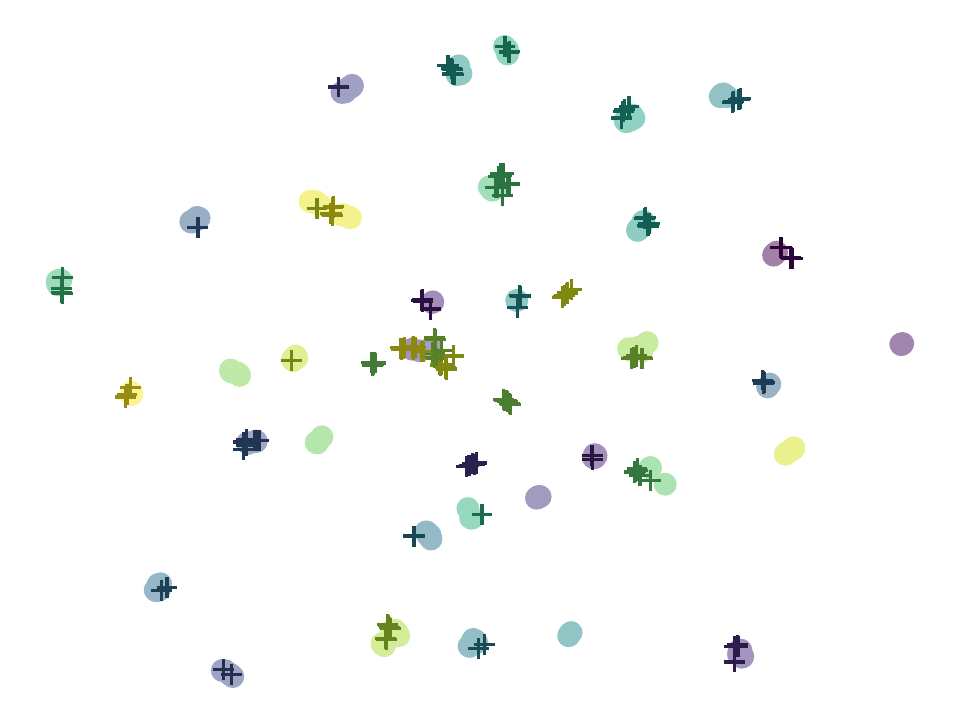}
        \caption{Sim-to-real \NaI{} scenario, ADDA model.}
        \label{fig:umap_nai_ADDA}
    \end{subfigure}

    \vspace{0.5em}
    \begin{subfigure}[t]{0.38\textwidth}
        \centering
        \includegraphics[width=\textwidth]{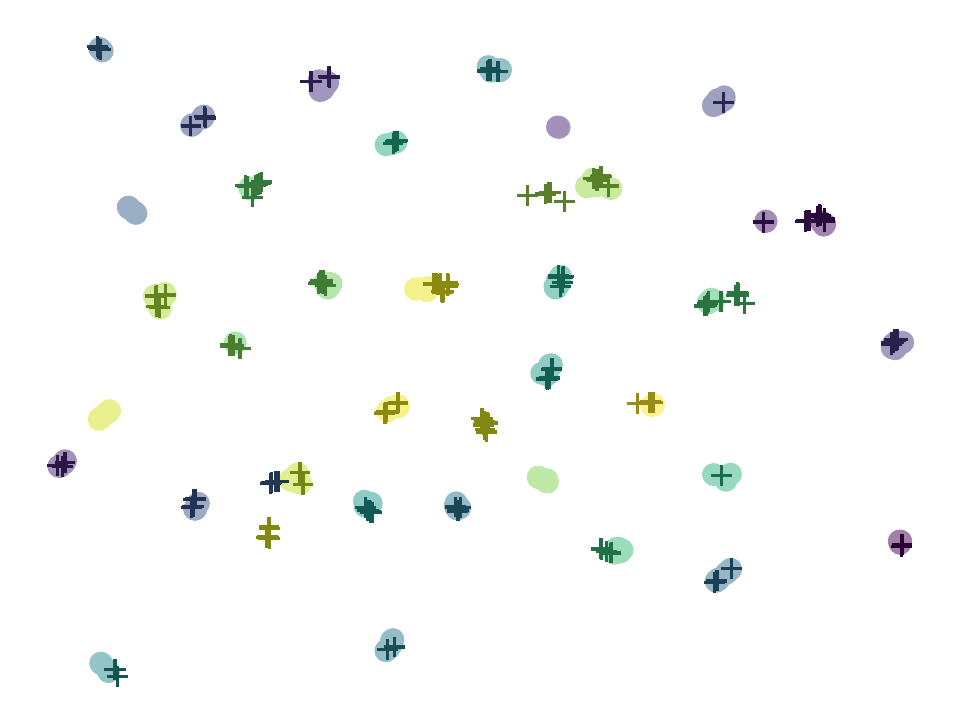}
        \caption{Sim-to-real \NaI{} scenario, DAN model.}
        \label{fig:umap_nai_DAN}
    \end{subfigure}
    \hfill
    \begin{subfigure}[t]{0.38\textwidth}
        \centering
        \includegraphics[width=\textwidth]{UDA_umap_TBNN_linear_gap_DANN_NaI.pdf}
        \caption{Sim-to-real \NaI{} scenario, DANN model.}
        \label{fig:umap_nai_DANN}
    \end{subfigure}

    \vspace{0.5em}
    \begin{subfigure}[t]{0.38\textwidth}
        \centering
        \includegraphics[width=\textwidth]{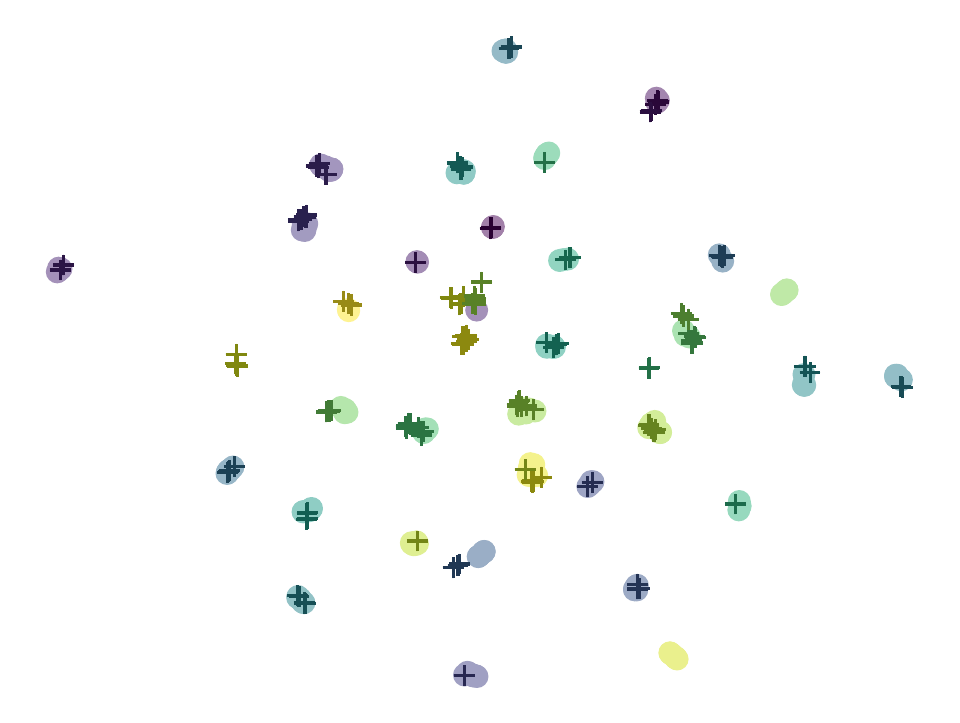}
        \caption{Sim-to-real \NaI{} scenario, DeepCORAL model.}
        \label{fig:umap_nai_DeepCORAL}
    \end{subfigure}
    \hfill
    \begin{subfigure}[t]{0.38\textwidth}
        \centering
        \includegraphics[width=\textwidth]{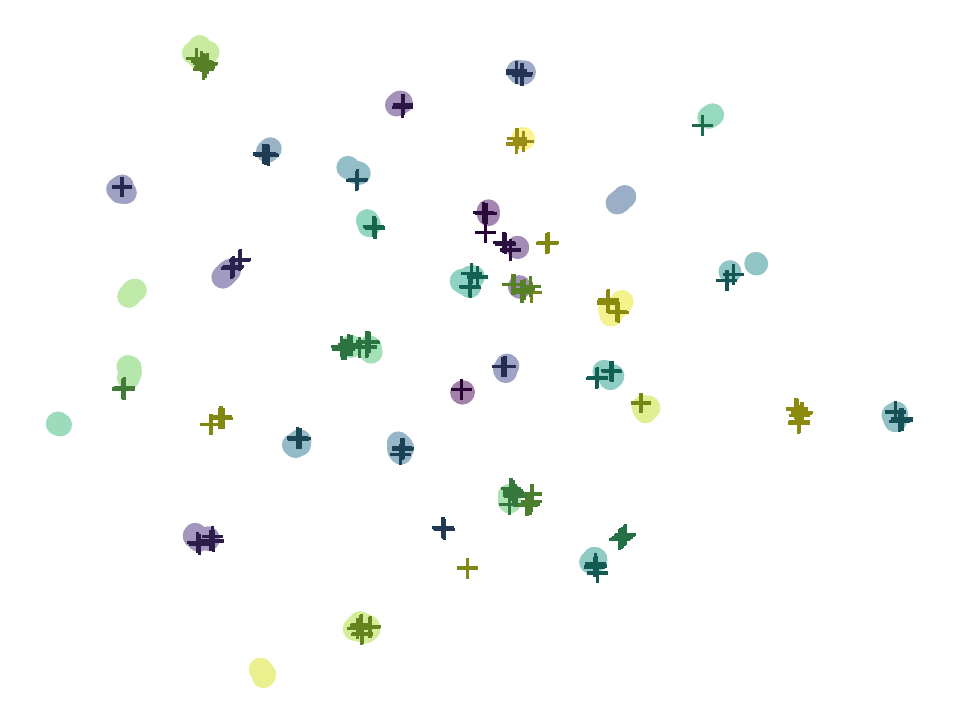}
        \caption{Sim-to-real \NaI{} scenario, DeepJDOT model.}
        \label{fig:umap_nai_DeepJDOT}
    \end{subfigure}

    \vspace{0.5em}
    \begin{subfigure}[t]{0.38\textwidth}
        \centering
        \includegraphics[width=\textwidth]{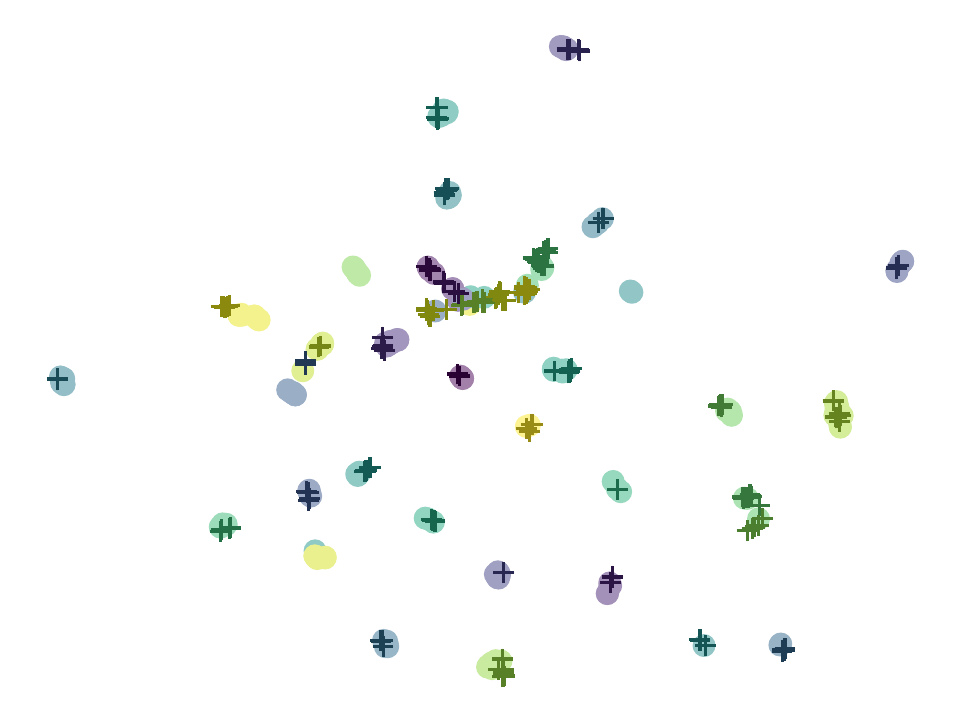}
        \caption{Sim-to-real \NaI{} scenario, Mean Teacher model.}
        \label{fig:umap_nai_MeanTeacher}
    \end{subfigure}
    \hfill
    \begin{subfigure}[t]{0.38\textwidth}
        \centering
        \includegraphics[width=\textwidth]{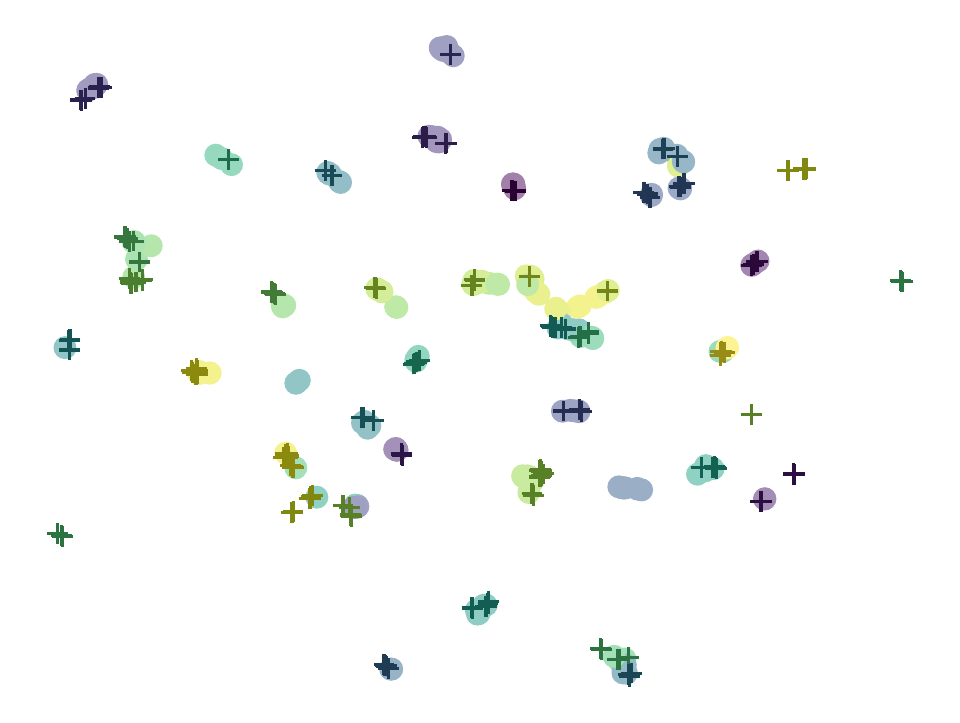}
        \caption{Sim-to-real \NaI{} scenario, SimCLR model.}
        \label{fig:umap_nai_SimCLR}
    \end{subfigure}

    \caption{Same as Fig.~\ref{fig:umap_hpge_all}, but for the sim-to-real \NaI{} domain adaptation scenario. Color indicates class.}
    \label{fig:umap_nai_all}
\end{figure}

\end{sloppypar}

\end{document}